%% file: main.tex
\documentclass[10pt,twocolumn,letterpaper]{article}
\usepackage{cvpr}
\usepackage{placeins}
\usepackage{subcaption}
\usepackage{tikz}
\usetikzlibrary{positioning}
\usepackage{pgfplots}
\usepackage{tabularx}
\usepackage{textcomp}
\usepackage{microtype}
\usepackage{colortbl}

\usepackage{amssymb}
\usepackage{pifont}
\newcommand{\cmark}{\ding{51}}
\newcommand{\xmark}{\ding{55}}

\input{preamble}

\definecolor{cvprblue}{rgb}{0.21,0.49,0.74}
\usepackage[pagebackref,breaklinks,colorlinks,citecolor=cvprblue]{hyperref}
\usepackage{multirow}

\title{StegoGAN: Leveraging Steganography\\ for Non-Bijective Image-to-Image Translation}

\author{Sidi Wu \textsuperscript{1}\thanks{Equal contribution.}
\qquad
Yizi Chen \textsuperscript{1}\footnotemark[1]
\qquad
Samuel Mermet \textsuperscript{2}
\qquad
Lorenz Hurni \textsuperscript{1}
\\
Konrad Schindler \textsuperscript{1}
\qquad
Nicolas Gonthier \textsuperscript{3,2}
\qquad
Loic Landrieu \textsuperscript{4,2}
\vspace{.5em}
\\
{\textsuperscript{1} \small  ETH Zurich
}
\quad
{
\textsuperscript{2} \small  Univ Gustave Eiffel, IGN, ENSG, LASTIG 
}
\quad
{
\textsuperscript{3} \small  IGN
}
\quad
{
\textsuperscript{4} \small Univ Gustave Eiffel, CNRS, Ecole des Ponts, LIGM
}
}

\begin{document}

\maketitle
\input{sec/0_abstract}    
\input{sec/1_intro}
\input{sec/2_relatedwork}

\input{sec/3_method}

\input{sec/4_experiment}

\input{sec/5_conclusion}

\pagebreak
\FloatBarrier
{
    \small
    \bibliographystyle{ieeenat_fullname}
    \bibliography{main}
}

\input{sec/X_suppl}

\end{document}

%% file: preamble.tex
\usepackage{mathtools}
\usepackage{tabularx,ragged2e,booktabs}
\usepackage{xcolor}

\newcommand{\eqdef}{\vcentcolon=}

\newcommand{\xgen}{x_{\text{gen}}}
\newcommand{\ygen}{y_{\text{gen}}}
\newcommand{\zgen}{z_{\text{gen}}}
\newcommand{\xrec}{x_{\text{rec}}}
\newcommand{\yrec}{y_{\text{rec}}}
\newcommand{\zrec}{z_{\text{rec}}}

\newcommand{\GXY}{G_{\cX \mapsto \cY}}
\newcommand{\GYX}{G_{\cY \mapsto \cX}}

\newcommand{\enc}{^\text{enc}}
\newcommand{\dec}{^\text{dec}}
\newcommand{\clean}{^\text{clean}}

\newcommand{\match}{^\text{match}}
\newcommand{\unmatch}{^\text{unmatch}}

\newcommand{\cX}{\mathcal{X}}
\newcommand{\cY}{\mathcal{Y}}
\newcommand{\cK}{\mathcal{K}}

\DeclareMathOperator{\upsample}{upsample}

%% file: sec/0_abstract.tex
\begin{abstract}
Most image-to-image translation models postulate that a unique correspondence exists between the semantic classes of the source and target domains. However, this assumption does not always hold in real-world scenarios due to divergent distributions, different class sets, and asymmetrical information representation. 
As conventional GANs attempt to generate images that match the distribution of the target domain, they may hallucinate spurious instances of classes absent from the source domain, thereby diminishing the usefulness and reliability of translated images. 
CycleGAN-based methods are also known to hide the mismatched information in the generated images to bypass cycle consistency objectives, a process known as steganography.
In response to the challenge of non-bijective image translation, we introduce StegoGAN, a novel model that leverages steganography to prevent spurious features in generated images. Our approach enhances the semantic consistency of the translated images without requiring additional postprocessing or supervision.
Our experimental evaluations demonstrate that StegoGAN outperforms existing GAN-based models across various non-bijective image-to-image translation tasks, both qualitatively and quantitatively. Our code and pretrained models are accessible at \url{https://github.com/sian-wusidi/StegoGAN}.
\end{abstract}

%% file: sec/1_intro.tex
\section{Introduction}
\label{sec:intro}
Image-to-image translation is an active research subject with impactful applications ranging from changing the style of images~\cite{CycleGAN2017, Gatys_2016_CVPR} to automatically creating maps from satellite images \cite{CycleGAN2017} or changing the modality of medical images \cite{dalmaz2022resvit}. When the source and target domains exhibit substantial differences, ensuring the semantic consistency between input images and their translation becomes particularly challenging~\cite{cohen2018distribution}. Our work explores the surprisingly uncharted field of adversarial, non-bijective image-to-image translation.

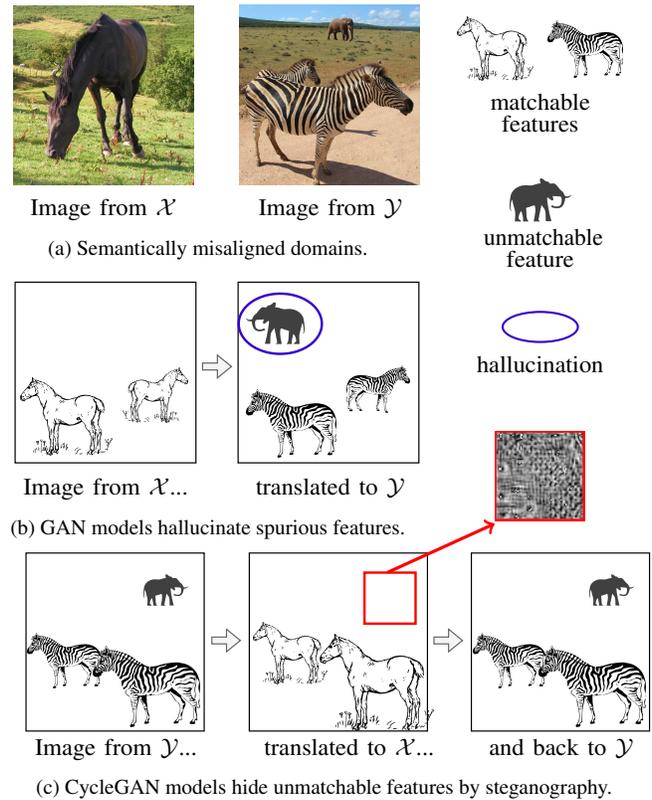
\begin{figure}
   \input{figures/teaser_bk4}
    \caption{{\bf Non-Bijective Translation.} 
    When image domains present classes without equivalence \subref{fig:teaser:a}, GAN models tend to hallucinate spurious features when translating images \subref{fig:teaser:b}. A related phenomenon is steganography, where CycleGAN-based models covertly encode features in low-amplitude patterns to bypass cycle consistency \subref{fig:teaser:c}. Instead of disabling this phenomenon, we harness steganography to prevent the hallucination of spurious features.}
    \label{fig:teaser}
\end{figure}

\noindent\textbf{Non-Bijective Image Translation.}
Existing translation methods assume a one-to-one correspondence between classes of the source and target domains: horses to zebras~\cite{CycleGAN2017}, satellite image features to their cartographic representation~\cite{CycleGAN2017, christophe2022neural}, or distinct cell types viewed under varying medical imaging modalities~\cite{dalmaz2022resvit}. 
However, as illustrated in~\Cref{fig:teaser:a}, this assumption does not always hold.
For instance, when considering a dataset with horses and one with zebras in their habitat, zebra images may include background elements with no equivalent in the source domain---like elephants.
Similarly, in map translation toponyms (\ie, place names printed onto the map) do not have counterparts in satellite images~\cite{christophe2022neural} . 
We qualify classes of the target domain without equivalent in the source domain as \emph{unmatchable}.

As illustrated in~\Cref{fig:teaser:b}, by trying to reproduce the distribution of the target domain, GANs may hallucinate \emph{spurious} features or textures in the generated images, \ie, objects without equivalent in the source image. This is particularly frequent for unmatchable classes~\cite{cohen2018distribution}. While this can be perfectly acceptable for some applications~\cite{li2019asymmetric}, adding nonexistent tumors in MRI scans or incorrect toponyms in maps can severely degrade the usefulness of the translation result.
Instead of detecting and removing these artifacts in post-processing, we propose an approach that directly prevents their generation using steganography.

\vspace{0.1cm}
\noindent\textbf{GAN Steganography.}  
To ensure its semantic consistency, CycleGAN~\cite{CycleGAN2017} back-translates the generated image to the source image.
However, unmatchable classes of the source domain cannot be encoded into meaningful features in the images generated in the target domain. As shown in~\Cref{fig:teaser:c}, these models can instead \emph{cheat} by encoding the necessary information into quasi-invisible patterns in the generated images~\cite{chu2017cycleganstego}. 
This process, known as steganography, allows GANs to perform seemingly impossible back-translation. For instance, in a map-to-satellite task, the model can restore the correct names of towns from satellite images that appear visually correct. This phenomenon is often viewed as a quirky optimization flaw, easily fixable by adding noise or blur~\cite{FuCVPR19-GcGAN, park2020cut, jung2022src}.

\vspace{0.1cm}\noindent\textbf{StegoGAN.}
We propose StegoGAN, a model that leverages steganography to detect and mitigate semantic misalignment between domains. 
In settings where the domain mapping is non-bijective, StegoGAN experimentally demonstrates superior semantic consistency over other GAN-based models both visually and quantitatively, without requiring detection or inpainting steps. 
In addition, we publish three datasets from open-access sources as a benchmark for evaluating non-bijective image translation models.

%% file: figures/teaser_bk4.tex
 \definecolor{mygreen}{rgb}{0.38, 0.66, 0.09}
    \definecolor{myorange}{rgb}{0.94, 0.64, 0.10}
    \definecolor{myblue}{rgb}{0.49, 0.65, 0.88}
    \definecolor{myfushia}{rgb}{1, 0.0, 1}
        \definecolor{myindigo}{rgb}{.22, 0.0, .8}

    \centering
    \hspace{-6.8mm}
    \begin{tabular}{l@{}l}
    
    \begin{subfigure}{.65\linewidth}
        \begin{tikzpicture}
        \node (imga) at (0,0) {
        \includegraphics[width=\linewidth]{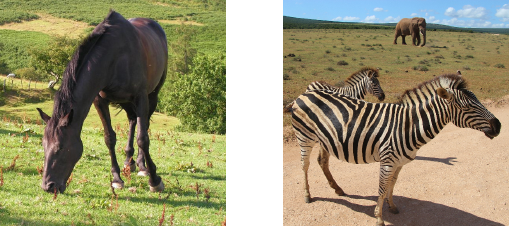}};
        \node [below left = -1mm and 3mm of imga.south,text width=.4\linewidth,align=center]{\small{Image from $\cX$ }};
        \node [below right = -1mm and 3mm of imga.south,text width=.4\linewidth,align=center]{\small{Image from $\cY$}}
        ;
        \end{tikzpicture}
        \caption{Semantically misaligned domains.}
        \label{fig:teaser:a}
    \end{subfigure}
    &
    \multirow[b]{2}{*}{
    \begin{minipage}[t][5cm][t]{.35\linewidth}
    \vspace{-3.8cm}
\begin{tikzpicture}[remember picture]
        
         \node at (-0.6,1.2) {\includegraphics[width=10mm]{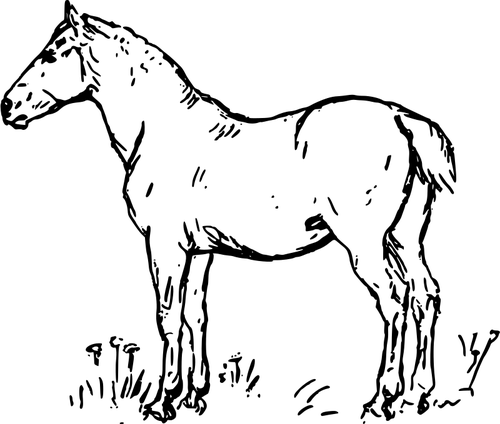}};
         \node at (0.6,1.2) {\includegraphics[width=10mm]{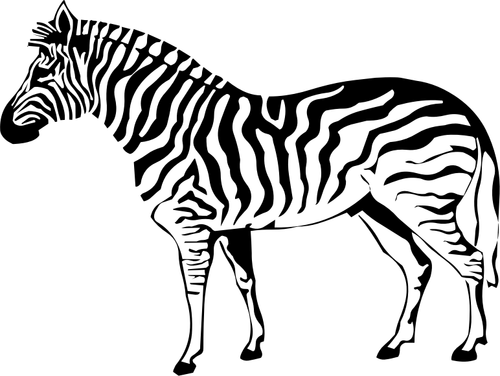}};
         
        \node [text width=1.5cm,align=center] at (0,0.5) {\small{matchable}};
        \node [text width=1.5cm,align=center] at (0,0.2) {\small{features}};

        \node at (0,-0.8) {\includegraphics[width=8mm]{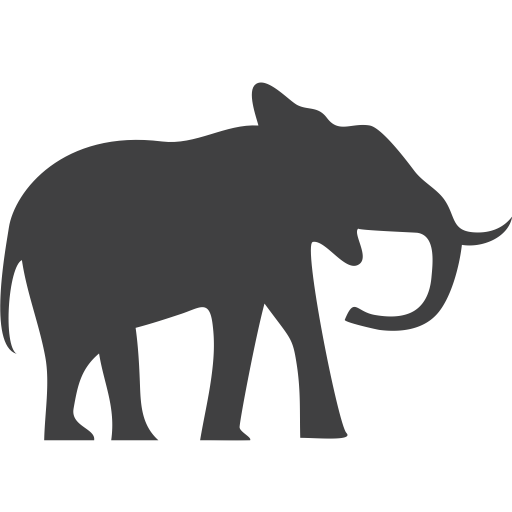}};
        
        \node [text width=1.5cm,align=center] at (0,-1.3) {\small{unmatchable}};
        \node [text width=1.5cm,align=center] at (0,-1.6) {\small{feature}};
        
        \draw [myindigo, thick] (0,-2.5) ellipse (5mm and 2mm);
        \node [text width=3cm,align=center] at (0,-3) {\small{hallucination}};

        \node[inner sep=0] at (0,-4.5) (steg){\includegraphics[width=12mm]{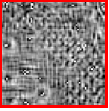}};
        \end{tikzpicture}

        \end{minipage}
}
    \\
    \begin{subfigure}{.65\linewidth}
        \begin{tikzpicture}
        \node (imga) at (0,0) {
        \includegraphics[width=\linewidth]{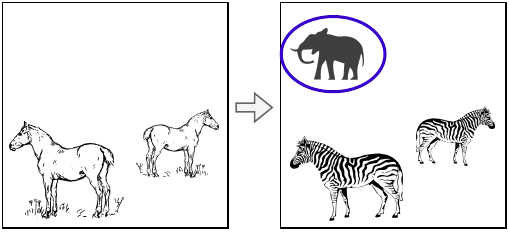}};
        \node [below left = -1mm and 3mm of imga.south,text width=.4\linewidth,align=center]{\small{Image from $\cX$... }};
        \node [below right = -1mm and 3mm of imga.south,text width=.4\linewidth,align=center]{\small{translated to $\cY$}}
        ;        
    \end{tikzpicture}
    \caption{GAN models hallucinate spurious features.}
            \label{fig:teaser:b}
    \end{subfigure}
    &
         \\
    \multicolumn{2}{c}{     
    \begin{subfigure}{\linewidth}
    \begin{tikzpicture}[remember picture]
     \node[inner sep=0] (imgb) at (0,0) {
         \includegraphics[width=\linewidth]{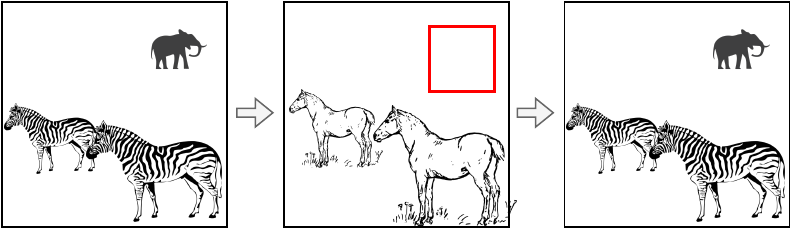}};
     \node [below left = -1mm and 16mm of imgb.south,text width=.3\linewidth,align=center]{\small{Image from $\cY$...}};
    \node [below left = -1mm and -15mm of imgb.south,text width=.3\linewidth,align=center]{\small{translated to $\cX$...}};   
    \node [below right = -1mm and 16mm of imgb.south,text width=.3\linewidth,align=center]{\small{and back to $\cY$}};
    \end{tikzpicture}
   
    \caption{CycleGAN models hide unmatchable features by steganography.}
    \label{fig:teaser:c}
    \end{subfigure}
    }
    \end{tabular}

    \begin{tikzpicture}[overlay, remember picture]
       \draw[red, very thick, ->] ([shift={(6.6mm, -2.9mm)}]imgb.north)--([shift={(0mm, 0mm)}]steg.south west);
    \end{tikzpicture}

%% file: sec/2_relatedwork.tex
\section{Related Work}
\label{sec:relatedworks}

\vspace{0.1cm}\noindent\textbf{GAN-Based Image Translation.}
GAN-based image translation models transfer the style of images between domains with an adversarial perceptual loss~\cite{goodfellow2014generative}. 
When pairs of aligned images from both domains are available, the translated images can also be supervised by their fidelity with target images~\cite{isola2017pix}.
In practice, such pairs are not always available or even possible to obtain. 
In the absence of explicit equivalence between images, preserving the semantics of the input in the generated image is a priority.
Multiple approaches have been proposed to address this challenge, such as density-based regularization~\cite{xie2022decent}, spatial mutual information~\cite{park2020cut, Wang2021InstancewiseHN, jung2022src}, or cycle consistency losses \cite{CycleGAN2017, Lee2018DRIT, huang2018munit}.

\vspace{0.1cm}\noindent\textbf{Asymmetric Image Translation.} 
Translating between domains with different semantic distributions is challenging. 
Existing approaches include focusing the network's attention on the most discriminative part of the input image~\cite{tang2021attentiongan}, augmenting the consistency loss with geometric transformations~\cite{FuCVPR19-GcGAN}, replacing the consistency reconstruction term with a contrastive loss~\cite{park2020cut, jung2022src}, or ensuring that the translation is robust to small perturbations of the input~\cite{jia2021srunit}.
However, these methods assume a bijective relationship between the classes of the source and target domains.

Closest to our work is the model of Li \etal~\cite{li2019asymmetric}, which uses an auxiliary variable to model the information loss from information-rich domains (such as natural images) to information-poor domains (such as label maps).
In turn, they use this variable to create realistic poor-to-rich domain translations. 
Our work differs as we precisely want to avoid the creation of spurious---albeit realistic---details when translating to a domain with unmatchable classes.

\vspace{0.1cm}\noindent\textbf{CycleGAN Steganography.}
Chu \etal ~\cite{chu2017cycleganstego} discovered that, when faced with unmatchable classes, CycleGAN~\cite{CycleGAN2017} hides information in low-amplitude and high-frequency signals.
The model uses these visually imperceptible patterns to recreate the source image and bypass the cycle loss.
This contradicts the intention of the cycle consistency loss and makes the model more vulnerable to adversarial attacks~\cite{chu2017cycleganstego}.
Luckily, multiple approaches can prevent steganography, such as blurring~\cite{FuCVPR19-GcGAN}, compressing~\cite{Dziugaite2016ASO}, or adding noise~\cite{jung2022src} to the generated source images in the back-translation. Alternatively, the back-translation from poor to rich domains can be omitted in the cycle consistency loss~\cite{sestini2023fun, YiLLR20portrait}.

While steganography in CycleGAN can be problematic, it also offers an opportunity to analyse distribution differences. In StegAnomaly \cite{baur2020steganomaly}, a model is trained to translate healthy brain scans into a low-entropy domain with cycle consistency. When removing high-frequency components, the model error reveals anomalous structures. This approach, like ours, harnesses steganography for insightful domain analysis, albeit with a different goal.

%% file: sec/3_method.tex
\section{Methods}
\label{sec:methods}
\mathchardef\mhyphen="2D

\input{figures/overall.tex}

We consider two image domains $\cX$ and $\cY$ with respective semantic class sets $\cK_\cX$ and $\cK_\cY$.
The domains $\cX$ and $\cY$ are considered bijective if there exists a function $\phi$ from $\cK_\cX$ to $\cK_\cY$ such that each class $k_\cX$ has a unique and natural semantically equivalent class $\phi(k_\cX)$ in $\cY$, and vice-versa. %
A class of $\cK_\cY$ is said to be \emph{unmatchable} if it doesn't have an equivalent in $\cK_\cX$.
While this notion is somewhat subjective, many applications have obvious examples: toponyms are unmatchable in satellite images, and tumors in scans of healthy patients.

\vspace{0.1cm}\noindent\textbf{Objective.} Our goal is to learn a mapping $G: \cX \mapsto \cY$ such that the translation $G(x)$ of any image $x \in \cX$ aligns stylistically with images from $\cY$, while preserving the semantic content of $x$.
If $\cK_\cY$ contains an unmatchable class $k\unmatch_\cY$, images translated from $\cX$ to $\cY$ should not contain any instances of $k\unmatch_\cY$.
However, in an attempt to match the distribution of $\cY$, GAN models often create spurious instances of unmatchable classes in their translated images. 
In this paper, we propose a method that employs steganography to prevent the generation of such spurious information.

\subsection{CycleGAN}
CycleGAN~\cite{CycleGAN2017} learns unpaired image translation by enforcing the consistency between the input image and its back-translation from the generated image.
It uses two generators $\GXY: \cX \mapsto \cY$ and $\GYX: \cY \mapsto \cX$, and two domain discriminators $D_\cX: \cX \mapsto \{0,1\}$, $D_\cY: \cY \mapsto \{0,1\}$ which predict whether a sample is generated (0) or real (1).

In the following, when considering images $x\in \cX$ or $y\in \cY$, we define the following short-hands: $\xgen \eqdef \GYX(y)$ and $\ygen \eqdef \GXY(x)$ for the generated images, and 
$\xrec \eqdef \GYX(\ygen)$ and 
$\yrec \eqdef \GXY(\xgen)$ for the reconstructed images. We now detail the losses of CycleGAN.

\vspace{0.1cm}\noindent\textbf{Adversarial loss.} 
The adversarial loss~\cite{goodfellow2014generative} 
encourages the discriminators $D_\cX$ and $D_\cY$ to distinguish between authentic and generated images, while pushing the generators $\GXY$ and $\GYX$ to create credible images:

\begin{align}\nonumber
\!\!\!\! \mathcal{L}_{\text{GAN}}
 &\!=\!
 \mathbb{E}_{y\sim \cY}
 \log(D_\cY(y))
\!+\!
\mathbb{E}_{x\sim \cX}
\log(1\!-\!D_\cY(\ygen)) \\   
 &\!+\!
 \mathbb{E}_{x\sim \cX}
 \log(D_\cX(x))
\!+\! 
\mathbb{E}_{y\sim \cY}
\log(1\!-\!D_\cX(\xgen)). 
    \label{eq:ganloss}
\end{align}

\vspace{0.1cm}\noindent\textbf{Cycle consistency.}
The cycle consistency loss ensures that the back-translation of $\ygen$ to domain $\cX$ is close to the original image $x$, and likewise for $\xgen$ and $y$:
\begin{align}
    \mathcal{L}_\text{cyc} = \mathbb{E}_{x\sim \cX}\Vert \xrec-x\Vert
    + \mathbb{E}_{y\sim \cY}\Vert \yrec-y\Vert~,
\label{eq:cycleloss}
\end{align}
with $\Vert \cdot \Vert$ the pixel-wise $L_1$ norm.%

\vspace{0.1cm}\noindent\textbf{Identity loss.} The identity loss regularizes the generators to be close to identity, generally improving color composition:
\begin{align}
    \!\!\!\mathcal{L}_\text{id} 
    \!=\!
    \mathbb{E}_{x\sim \cX}\Vert \GYX(x)\!-\!x\Vert
    \!+\!
    \mathbb{E}_{y\sim \cY}\Vert \GXY(y)\!-\!y\Vert.
\label{eq:idtloss}
\end{align}
\textbf{Final Loss.} The final objectives are:
\begin{align}\label{eq:loss:cycleGAN:gene}
    \mathcal{L}(\GXY,\GYX) 
    &
    = \mathcal{L}_{GAN} 
    + \lambda_\text{cyc}\mathcal{L}_\text{cyc} + \lambda_\text{id}\mathcal{L}_\text{id}~, \\\label{eq:loss:cycleGAN:disc}
    \mathcal{L}(D_{\cX},D_{\cY}) 
   &= -\mathcal{L}_\text{GAN}~,
\end{align}
with $\lambda_\text{cyc}$ and $\lambda_\text{id}$ non-negative hyperparameters. 

\subsection{StegoGAN}
We introduce StegoGAN, a novel model building on the CycleGAN framework~\cite{CycleGAN2017}, designed specifically for scenarios where domains $\cX$ and $\cY$ lack a bijective relationship. When generating images $\ygen$ in $\cY$, the generator $\GXY$ may add spurious instances of an unmatchable class $k\unmatch_\cY$ in order to deceive the discriminator $D_\cY$. Our goal is to prevent the generation of such hallucinated instances. 
To achieve this, we leverage steganography to explicitly disentangle the matchable and unmatchable information in the backward cycle ($y \mapsto \xgen \mapsto \yrec$) and prevent the network from hallucinating in the forward translation ($x \mapsto \ygen \mapsto \xrec$).
See~\Cref{fig:overall} for the overall design of our approach.

\vspace{0.1cm}\noindent\textbf{Steganography.} 
In order to faithfully reconstruct $y$ with $\yrec$, the translated image $\xgen$ must somehow contain information about the instances of the unmatchable class $k\unmatch_\cY$. CycleGAN methods typically achieve this by hiding low-amplitude and high-frequency patterns in $\xgen$ that will be decoded by $\GXY$ and translated back to instances of $k\unmatch_\cY$ in $\yrec$.
Adding low-intensity noise on each pixel is typically sufficient to destroy the hidden information and prevent steganography~\cite{chu2017cycleganstego}.

Steganography is often viewed as an optimization flaw that undermines the consistency loss. However, in the case of non-bijective translation, this is the only way to let $\GXY$ reconstruct instances of $k\unmatch_\cY$ in $\yrec$. 
Instead of disabling steganography, we propose to use it to our advantage to detect and prevent spurious generations. 
We adapt CycleGAN so that steganography takes place in 
 feature-space instead of pixel-space, and in an explicit manner.

\vspace{0.1cm}\noindent\textbf{Backward Cycle.} We decompose the generators $\GXY$ and  $\GYX$ into two components: an encoder and a decoder, such that $\GXY=\GXY\dec \circ  \GXY\enc$ and $\GYX=\GYX\dec \circ \GYX\enc$. The encoders map their inputs to feature maps of spatial dimension $H \times W$, where each pixel has $C$ channels. In the following, we denote the intermediary representation of $y$ in $\GYX$ by $\zgen\eqdef \GYX\enc(y)$. The feature map $\zgen$ encodes information about both matchable and unmatchable classes, which we want to disentangle.

We introduce a network $M$: $\mathbb{R}^{H \times W \times C} \mapsto [0,1]^{H \times W \times C}$ that assigns an \emph{unmatchability} score between $0$ and $1$ to each pixel and channel. Here, a score of $1$ indicates that the information does not have a counterpart in domain $\cX$, while it does for $0$. 
This process gives us the {unmatchability mask} $M(\zgen)$, which we use to split $\zgen$ into its matchable and unmatchable parts:
\begin{align}
    \zgen\unmatch &= M(\zgen) \odot z \\
    \zgen\match &= \left(1-M(\zgen)\right) \odot z~,
    \label{eq:matchable}
\end{align}
with $\odot$ the pixel-wise and channel-wise Hadamard product.
In our model, the  generated image $\xgen$ is computed using only the matchable part of the representation:
\begin{align}
    \xgen = \GYX\dec(\zgen\match)~.
\end{align}
We produce two reconstructions of $y$: $\yrec\clean$, which is a direct back-translation of $\xgen$ into the $\cY$ domain : $\yrec\clean = \GXY (\xgen)$; and $\yrec$, which is generated by decoding a combination of the unmatchable part of $\zgen$ and the features extracted by $\GXY\enc$ from a noise-perturbed version of $\xgen$:
\begin{align}
    \yrec = 
    \GXY\dec\left(
        \GXY\enc
        \left(
            \xgen 
            +
            \epsilon
        \right)
        +
        \zgen\unmatch
    \right)~,
\end{align}
with $\epsilon$ denoting random Gaussian noise of low amplitude applied to each pixel and channel of $\xgen$. This noise is added to destroy potential steganographic information in $\xgen$, therefore forcing $\GXY$ to rely only on $\zgen\unmatch$ to reconstruct unmatchable features in $\yrec$. 

The key mechanisms to disentangle matchable and unmatchable information are twofold: (i) disturbing direct steganography with random noise, and (ii) explicitly providing unmatchable information to $\GXY$ \emph{in feature-space}.

\vspace{0.1cm}\noindent\textbf{Forward Cycle.} In the forward cycle $x \rightarrow \ygen \rightarrow \xrec$, the generator $\GXY$ may create spurious instances of unmatchable classes when translating $x$ to $\cY$ to fulfill the expectations of the discriminator $D_{y}$. 
To address this, we perform two distinct translations of $x$ in $\cY$: $\ygen$ has explicit access to the steganographic information $\zgen\unmatch$ extracted from the backward cycle, while $\ygen\clean$ does not:
\begin{align}
    &\ygen = \GXY\dec\left(\GXY\enc(x) + \zgen\unmatch \right)\\
    &\ygen\clean = \GXY\left(x\right)~.
\end{align}
The rationale is that $\GXY\dec$ has explicit access to information about the unmatchable classes of $y$, so it is not incentivized to invent them.
For consistency with the backward step, where the decoder of $\GYX$ processes only matchable information as defined in \eqref{eq:matchable}, we use the same disentanglement approach for generating $\xrec$:
\begin{align}
   \xrec = 
   \GYX\dec\left(
       \left(
            1-M\left(
                \zrec
                \right)
       \right)
       \odot
       \zrec
        \right)~,
\end{align}
where $\zrec=\GYX\enc(\ygen)$ is the intermediary representation of $\ygen$ in the forward cycle.

\input{figures/qualitative_comparison}

\vspace{0.1cm}\noindent\textbf{Mask Regularization.}
To avoid degenerate behaviors of our explicit steganography mechanism, we enforce two priors on the unmatchability masks: (i) given that a well-posed translation problem predominantly involves matchable features, the masks should be sparse; (ii) to improve the model's interpretability, we favor mask values near $0$ or $1$, representing clear decisions about matchability. To enforce these priors, we regularize the masks with the non-convex $L_{0.5}$ norm~\cite{xu2010L0.5}:
\begin{equation}
    \mathcal{L}_{reg} = 
      \mathbb{E}_{y\sim \cY}
    \Vert M(\zgen)\Vert_{0.5}
    + 
     \mathbb{E}_{x\sim \cX}
    \Vert M(\zrec)\Vert_{0.5}~.
\end{equation}

\vspace{0.1cm}\noindent\textbf{Matchable Consistency.}
The images $\ygen$ and $\ygen\clean$, as well as  $\yrec$ and $\yrec\clean$, should be identical outside of unmatchable regions. 
To enforce this constraint, we design a function $I$ which takes an unmatchability mask $m$ as input, takes the channel-wise maximum values of all pixels, flips its value from $[0,1]$ to $[1,0]$, and upsamples the results to the dimensions of the input images.
\begin{align}
    I(m) = \upsample \left( 1-\max_c m\right)~.
\end{align}
The obtained consistency masks $I(m)$ have values close to $1$ for pixels with only matchable content, and close to $0$ otherwise. This enables us to define a loss for $\ygen\clean$ and $\yrec\clean$ that focuses solely on regions with matchable features:
\begin{align}\nonumber
  \mathcal{L}_\text{match} &=  
  \mathbb{E}_{y\sim \cY}
    \Vert
    I\left(M(\zgen)\right) \odot 
    \left( 
        \ygen  - \ygen\clean 
    \right)
    \Vert
    \\
    &+
    \mathbb{E}_{x\sim \cX}
     \Vert
    I\left(M(\zrec)\right) \odot 
    \left( 
        \yrec  - \yrec\clean
    \right)
    \Vert~.  
\end{align}

\vspace{0.1cm}\noindent\textbf{Final Objective.} 
In addition to the standard CycleGAN loss components (\ref{eq:loss:cycleGAN:gene}-\ref{eq:loss:cycleGAN:disc}), we integrate $\mathcal{L}_\text{match}$ and $\mathcal{L}_\text{reg}$ into the overall loss function $\mathcal{L}(\GXY,\GYX)$, weighted by their respective coefficients $\lambda_\text{reg}$ and $\lambda_\text{match}$. Crucially, our proposed approach remains unsupervised, requiring neither aligned images from $\cX$ and $\cY$ nor specific annotations of unmatchable features.

%% file: figures/overall.tex
\begin{figure*}
    \centering
    \begin{tabular}{c@{}c}
        \multicolumn{2}{c}{
        \hspace{-0.5cm}
        \includegraphics[width=\linewidth]{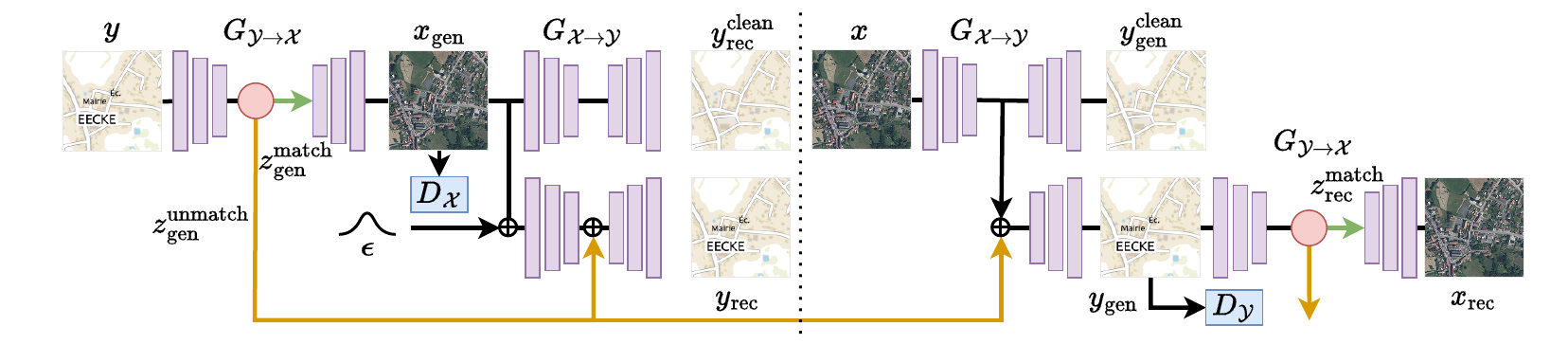}
        }
        \\
        \begin{subfigure}{.5\linewidth}
        \caption{Backward cycle}
        \label{fig:main:a}
        \end{subfigure} 
        &
        \begin{subfigure}{.5\linewidth}
        \caption{Forward cycle}
         \label{fig:main:b}
        \end{subfigure} 
        \\
        \multicolumn{2}{c}{
        \begin{tabular}{r@{}c@{}c}
        \begin{subfigure}{.2\linewidth}
        \raisebox{0cm}[0pt][0pt]{\centering
        \includegraphics[height=.1\textheight]{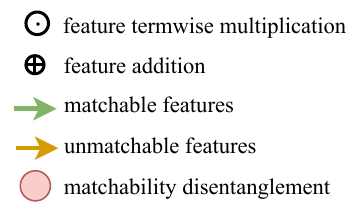}
        }
        \end{subfigure}
         &
         \begin{subfigure}{.5\linewidth}
         \centering
        \includegraphics[height=.1\textheight]{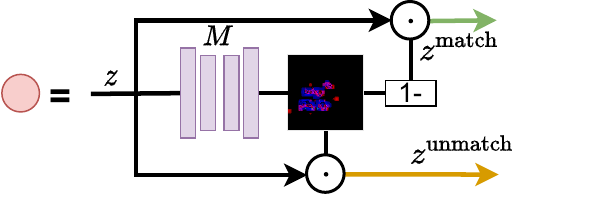}
         \caption{Matchability disentanglement module}
          \label{fig:main:c}
         \end{subfigure} 
        &
        \begin{subfigure}{.3\linewidth}
        \centering
       \includegraphics[height=.1\textheight]{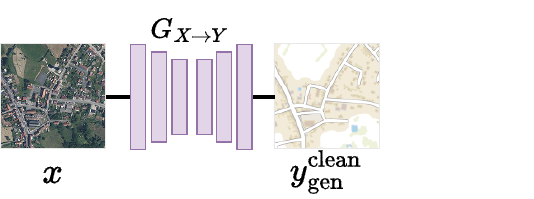}
    \caption{Inference}
     \label{fig:main:d}
    \end{subfigure} 
    \end{tabular}
    }
    \end{tabular}
        \vspace{-0.25cm}
    \caption{{\bf Architecture.}
    To avoid spurious generation of unmatchable classes in non-bijective image translation, we propose to make the steganographic process explicit and in feature-space. 
    Our model runs the backward cycle first  \subref{fig:main:a}, then the forward translation cycle \subref{fig:main:b}. 
    Thanks to our matchability disentanglement module \subref{fig:main:c},  we can separate the matchable and unmatchable information while translating images from domain $\cY$ to $\cX$. We can then produce generated and reconstructed images with and without unmatchable features. At inference time \subref{fig:main:d}, our model operates like a normal image translation model.}
    \vspace{-0.25cm}
    \label{fig:overall}
\end{figure*}

%% file: figures/qualitative_comparison.tex
\begin{figure*}[t]
  \centering
  \setlength\tabcolsep{1pt}
  \begin{tabularx}{\linewidth}{cccccccc}
    Source & Ground Truth & \bf StegoGAN & CycleGAN \cite{CycleGAN2017} & DRIT \cite{Lee2018DRIT} & GcGAN \cite{FuCVPR19-GcGAN} & CUT \cite{park2020cut} & SRUNIT \cite{jia2021srunit}\\
    \includegraphics[width=0.12\linewidth]{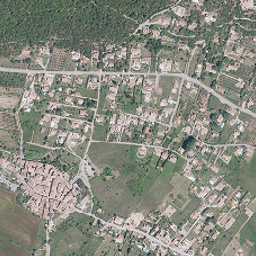}&
    \includegraphics[width=0.12\linewidth]{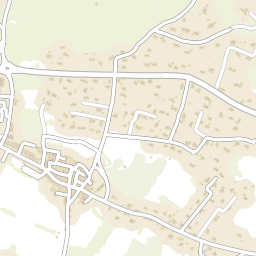}&
    \includegraphics[width=0.12\linewidth]{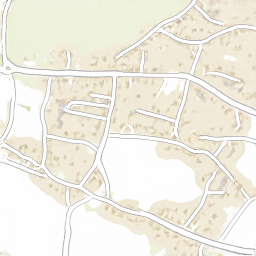}&
    \includegraphics[width=0.12\linewidth]{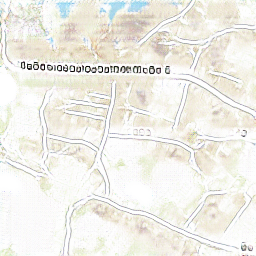}&
    \includegraphics[width=0.12\linewidth]{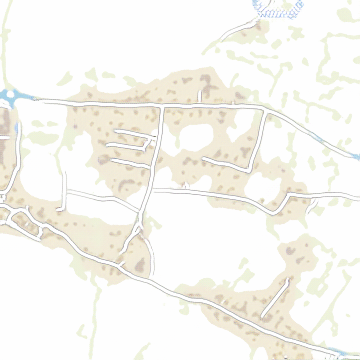}&
    \includegraphics[width=0.12\linewidth]{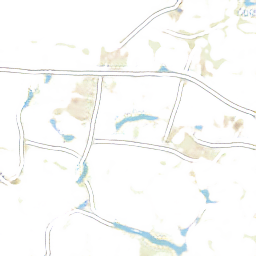}&
    \includegraphics[width=0.12\linewidth]{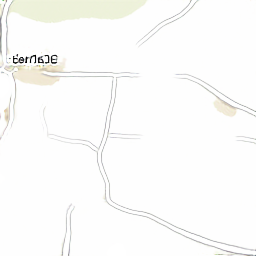}&
    \includegraphics[width=0.12\linewidth]{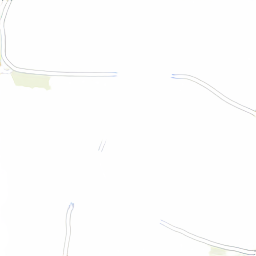}\\
    \includegraphics[width=0.12\linewidth]{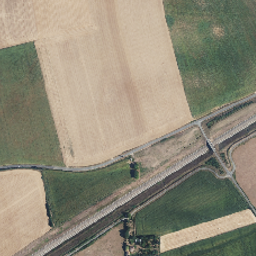}&
    \includegraphics[width=0.12\linewidth]{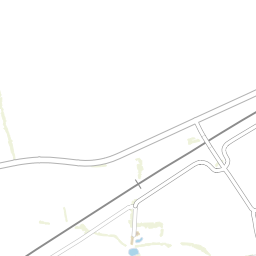}&
    \includegraphics[width=0.12\linewidth]{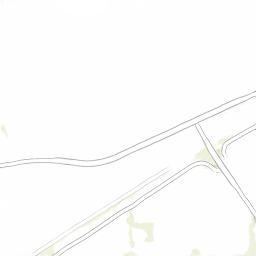}&
    \includegraphics[width=0.12\linewidth]{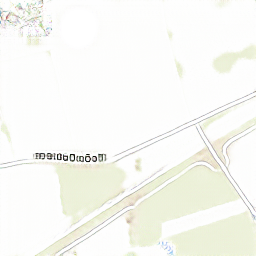}&
    \includegraphics[width=0.12\linewidth]{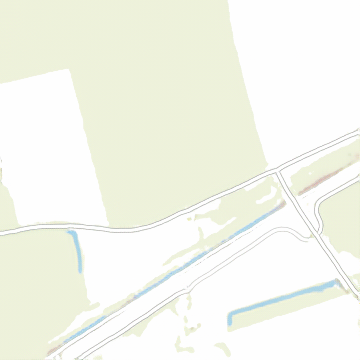}&
    \includegraphics[width=0.12\linewidth]{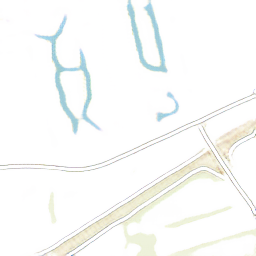}&
    \includegraphics[width=0.12\linewidth]{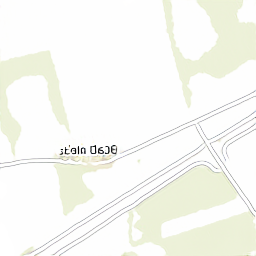}&
    \includegraphics[width=0.12\linewidth]{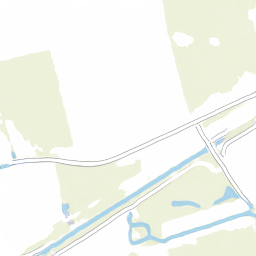}\\
    \includegraphics[width=0.12\linewidth]{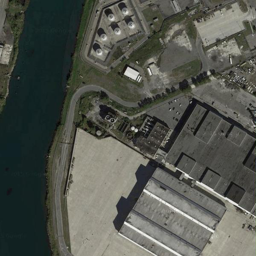}&
    \includegraphics[width=0.12\linewidth]{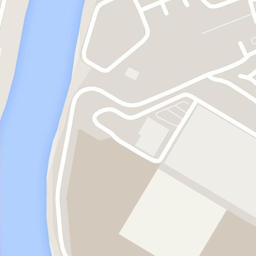}&
    \includegraphics[width=0.12\linewidth]{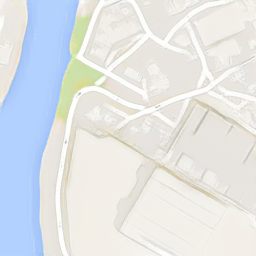}&
    \includegraphics[width=0.12\linewidth]{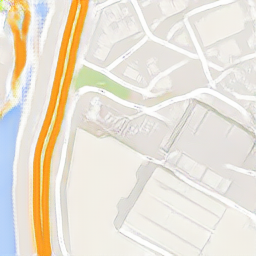}&
    \includegraphics[width=0.12\linewidth]{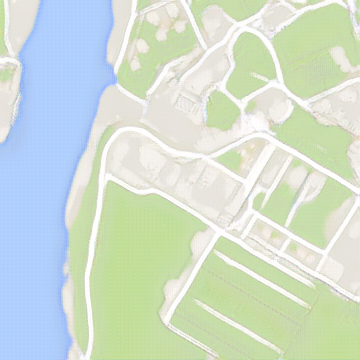}&
    \includegraphics[width=0.12\linewidth]{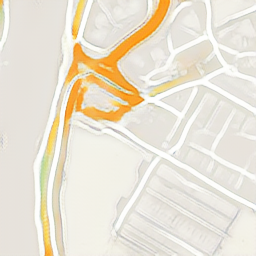}&
    \includegraphics[width=0.12\linewidth]{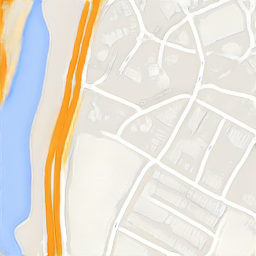} &
    \includegraphics[width=0.12\linewidth]{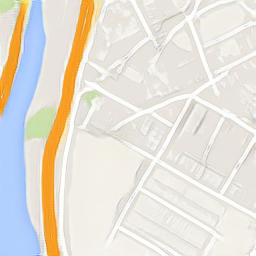}
    \\
    \includegraphics[width=0.12\linewidth]{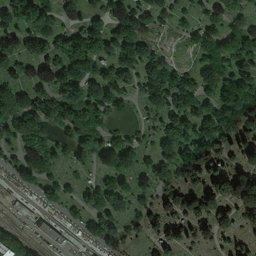}&
    \includegraphics[width=0.12\linewidth]{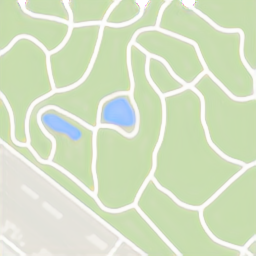}&
    \includegraphics[width=0.12\linewidth]{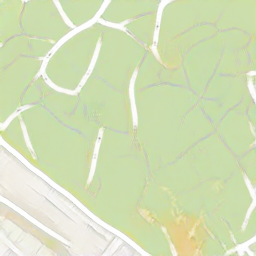}&
    \includegraphics[width=0.12\linewidth]{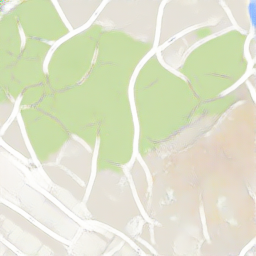}&
    \includegraphics[width=0.12\linewidth]{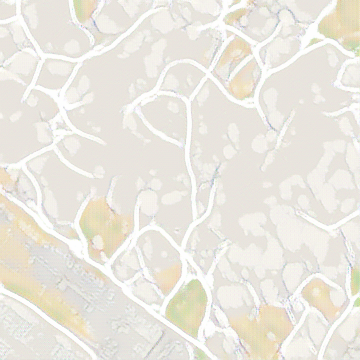}&
    \includegraphics[width=0.12\linewidth]{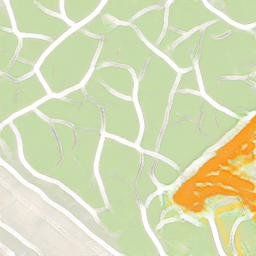}&
    \includegraphics[width=0.12\linewidth]{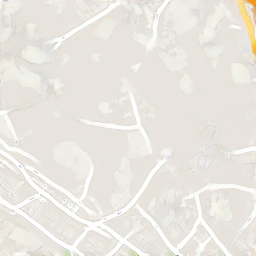} &
    \includegraphics[width=0.12\linewidth]{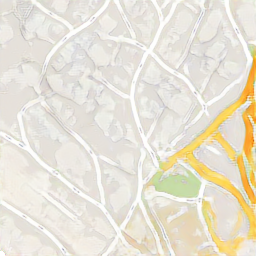} \\
    \includegraphics[angle=-0, width=0.12\linewidth]{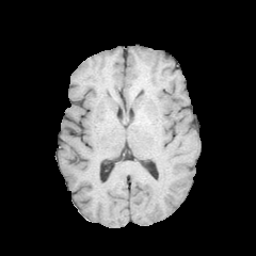}
    &
    \includegraphics[angle=-0, width=0.12\linewidth]{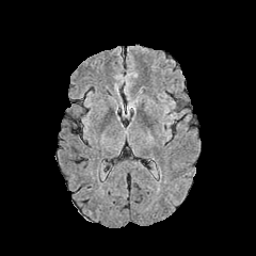}
    &
    \includegraphics[angle=-0, width=0.12\linewidth]{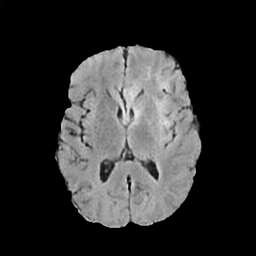}
    &
    \includegraphics[angle=-0, width=0.12\linewidth]{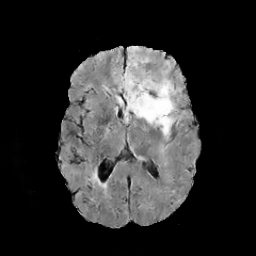}
    &
    \includegraphics[angle=-0, width=0.12\linewidth]{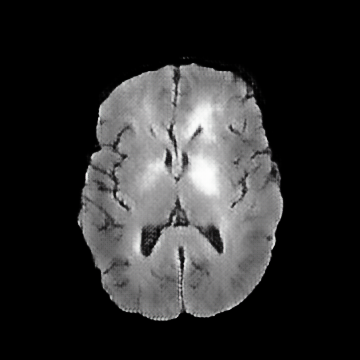}
    &
    \includegraphics[angle=-0, width=0.12\linewidth]{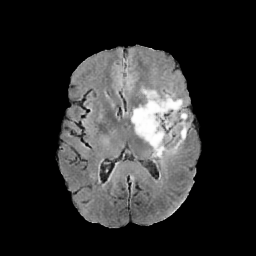}
    &
    \includegraphics[angle=-0, width=0.12\linewidth]{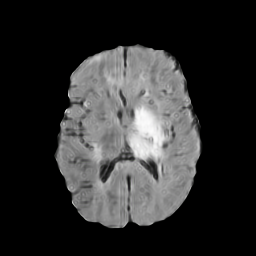}
    &
    \includegraphics[angle=-0, width=0.12\linewidth]{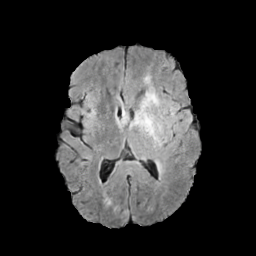}\\
    \includegraphics[angle=-0, width=0.12\linewidth]{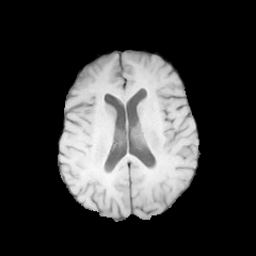}
    &
    \includegraphics[angle=-0, width=0.12\linewidth]{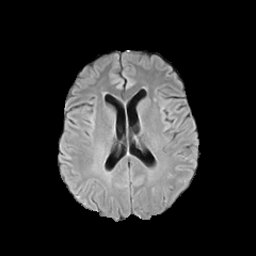}
    &
    \includegraphics[angle=-0, width=0.12\linewidth]{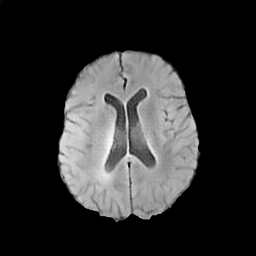}
    &
    \includegraphics[angle=-0, width=0.12\linewidth]{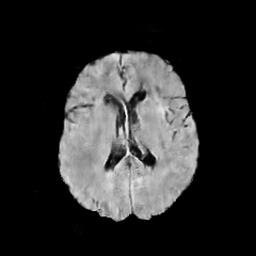}
    &
    \includegraphics[angle=-0, width=0.12\linewidth]{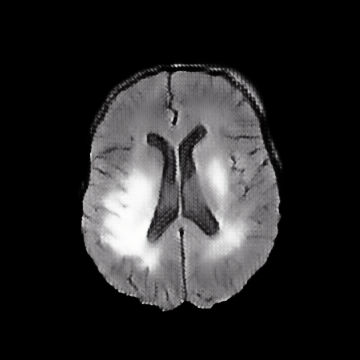}
    &
    \includegraphics[angle=-0, width=0.12\linewidth]{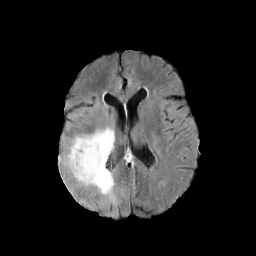}
    &
    \includegraphics[angle=-0, width=0.12\linewidth]{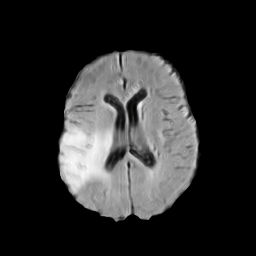}
    &
    \includegraphics[angle=-0, width=0.12\linewidth]{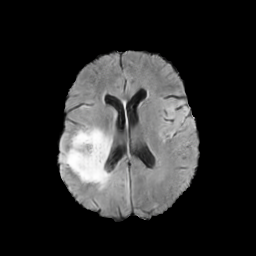}
  \end{tabularx}
  \vspace{-0.5em}
  \caption{{\bf Qualitative Comparison.} 
  We report reconstructions from the test sets of PlanIGN (top two rows), GoogleMap (row 3 and 4), and MRI (last two rows). 
  Contrary to the other models, StegoGAN does not hallucinate spurious toponyms, highways (orange roads), or tumors (white areas) and shows better semantic correspondences during translation.
  }
  \label{fig:all_qualitative}
\end{figure*}

%% file: sec/4_experiment.tex
\input{tex_table/results_on_google_map}
\section{Experiments}
\label{sec:experiments}
In this section, we assess the improvements brought by our method for non-bijective image translation across various datasets and compare them with existing models, both qualitatively and quantitatively.

\vspace{0.1cm}\noindent\textbf{Implementation details.}
We follow the setting of CyleGAN~\cite{CycleGAN2017} as our baseline model: 
the generators are Resnets~\cite{he2016res} and the discriminator is based on PatchGAN~\cite{isola2017pix}. 
We define the encoders as the first half of the generator's layers and the decoders as the second half.
The unmatchability mask predictor $M$ is defined as a small 3-layer convolutional neural network (CNN).
We set $\lambda_\text{cyc}=10$, $\lambda_\text{idt}=0.5$ as in~\cite{CycleGAN2017} and $\lambda_\text{match}=1$.
The amplitude of the perturbation $\epsilon$ is $0.01$ as in~\cite{chu2017cycleganstego}.
The Appendix provides more architecture and training details.

\vspace{-.25em}
\subsection{Datasets}
We assess the performance of StegoGAN across several image translation tasks that feature unmatchable classes. 
Each dataset follows a consistent structure: the training set includes images from the source domain $\cX$ devoid of a specific class (\eg, tumors, motorways, toponyms) while the target domain $\cY$ does include that class. 
The test set comprises \emph{paired} images from both domains, specifically excluding the unmatchable class. 
This setup allows us to quantify the models' hallucinations: any generated instances of the unmatchable class are necessarily spurious and due to its presence in the training set.
We release all three curated datasets on the \href{https://zenodo.org/records/10839841}{Zenodo platform} and provide details below.

\vspace{0.1cm}\noindent\textbf{\underline{PlanIGN.} $\cX:$ Aerial Photo, $\cY:$ Maps, $k\unmatch_\cY:$ Text.} 
We construct a dataset using open data from the French National Mapping Agency (IGN), comprising $1900$ aerial (ortho-)images at $3$m spatial resolution, and two versions of their corresponding topographic maps: one with and one without toponyms.
This dataset presents a clear unmatchable class: place names.
The training set includes $1000$ maps with toponyms and $1000$ aerial images, while the test set comprises $900$ map samples without toponyms and their corresponding aerial images.

\vspace{0.1cm}\noindent\textbf{\underline{GoogleMaps.} $\cX:$ Aerial Photo, $\cY:$ Maps, $k\unmatch_\cY:$ Highways.} 
The GoogleMaps dataset~\cite{isola2017pix} is a standard benchmark for image translation tasks~\cite{jung2022src,jia2021srunit}. 
It contains $1096$ map/image pairs for training and $1098$ for testing.
To create a controlled non-bijective scenario,
we exclude all satellite images that show highways and sample the maps of the training set to contain varying proportions of maps with highways, ranging from $0\%$ to $65\%$, for a fixed total of $548$ maps. 
For the test set we selected $898$ pairs without highways.

\vspace{0.1cm}\noindent\textbf{\underline{Brats MRI} $\cX:$ T1 Scans, $\cY:$ FLAIR, $k\unmatch_\cY:$ Tumors.} 
Lastly, we used a dataset of brain MRI scans~\cite{menze2014multimodal} with two modalities: T1 (naive) and FLAIR (T2 Fluid Attenuated Inversion Recovery)~\cite{hajnal1992use}.
We adapt the protocol that Cohen \etal~\cite{cohen2018distribution} used for the Brats2013 datasets~\cite{menze2014brats13} to the more recent Brats2018~\cite{bakas2018ibrats17} dataset by varying the percentage of scans with tumors in the target domain.
We selected transverse slices from the $60^\circ$ to $100^\circ$ range in the caudocranial direction~\cite{andermatt2019pathology} for both T1 and FLAIR scans. 
Each scan was classified as tumorous if more than $1\%$ of its pixels were labeled as such, and as healthy if it contained no tumor pixels. 
The training set contains $800$ images from each modality, with all source images (T1) being healthy and the target domain (FLAIR) comprising $60\%$ tumorous scans. 
The test set contains $335$ paired scans of healthy brains.

\subsection{Evaluation metrics}

We use a broad range of metrics to evaluate the performance of StegoGAN and other image translation algorithms in the non-bijective setting:

\vspace{0.1cm}\noindent\textbf{FID and KID.} The Fréchet Inception Distance (FID) 
~\cite{heusel2017fid} and Kernel Inception Distance (KID)~\cite{binkowski2018kid} are widely used to quantify the similarity between the distributions of real and generated images in the target domain.

\vspace{0.1cm}\noindent\textbf{RMSE, Acc($\sigma_1$), and Acc($\sigma_2$).} As the test sets comprise paired images from both domains, we can directly compare the Root Mean Square Error (RMSE) between the real and predicted images in the target domain. We count a predicted pixel as correctly predicted if it deviate by less than a fixed threshold in any of the color channels \cite{FuCVPR19-GcGAN}. We use $\sigma_1=5$ and $\sigma_2=10$ for the GoogleMap dataset and $\sigma_1=2$ and $\sigma_2=5$ for the less colorful PlanIGN dataset.

\vspace{0.1cm}\noindent\textbf{pFPR and iFPR.}
In the GoogleMap dataset, highways are always depicted in orange, allowing us to label pixels where all color channels differ by less than $20$ units from $(240,160,30)$ as highways. In the Brats MRI dataset, we use a pretrained tumor detector~\cite{buda2019tumorpretrain} to find spurious tumors in the generated images. This allows us to compute the average false positive rate per pixel (pFPR) and per instance (iFPR) of the generated images.

\subsection{Results}
\input{tex_table/PlanIGN_results.tex}

\textbf{Qualitative Results.} 
\Cref{fig:all_qualitative} showcases StegoGAN's qualitative performance against other image translation algorithms. Notably, StegoGAN effectively avoids generating unmatchable classes such as texts, highways, and tumors, while producing high-quality image translations.

\vspace{0.1cm}\noindent\textbf{Quantitative Results.} 
On the PlanIGN dataset (\Cref{tab:planIGN_results}) and the Brats MRI dataset (\Cref{tab:quantitativetumor}), StegoGAN outperforms others in fidelity, achieving the lowest RMSE by a margin of $4.5$ on PlanIGN and by $3.5$ for Brats MRI. 
Furthermore, it significantly enhances pixel accuracy, with improvements of $+11.6$ in Acc($\sigma_1$) and $+17.2$ in Acc($\sigma_2$) on PlanIGN. 
In the MRI dataset, StegoGAN dramatically reduces false positive rates—--over 20$\times$ lower than CycleGAN and 10$\times$ less than the next best model SRUNIT (for pFPR).

On the GoogleMap dataset, as shown in~\Cref{fig:tex_table/google_results_graph}, StegoGAN's performance is on par with CycleGAN at $0\%$ unmatchable cases and remains stable even as this ratio increases, unlike other methods that degrade. Remarkably, StegoGAN maintains a consistent false positive rate of $0$ across all tests, while this rate increases for all other methods.

\input{tex_table/Brats_result_0.6}
\input{figures/unmatchability_masks}

\vspace{0.1cm}\noindent\textbf{Unmatchability Masks.} 
In~\Cref{fig:visual_mask}, we illustrate the emergent ability of the unmatchability masks to trace the outline of unmatchable class instances like toponyms, highways, and tumors. This aspect highlights the versatility of our approach, which functions without explicit supervision or aligned images, offering a tool to explore the pairwise semantic differences between arbitrary datasets.
\subsection{Ablation study and analysis}\label{sec:ablation}
We explore the impact of our main design choices, as well as further capabilities and limitations of our approach. See the Appendix for further ablations.
\input{tex_table/PlanIGN_ablation_depth}

\input{figures/Ablation_mask}

\vspace{0.1cm}\noindent\textbf{Encoder Ablation.} 
We conducted an ablation study on the definition of the intermediary representation $z$ by varying the depth at which the "encoder" ends and the "decoder" starts within the generator. Given that our generators consist of $9$ consecutive convolutional blocks, we experimented with different configurations: $-1$ (indicating no decoder), $1$ (the configuration used in our paper), as well as depths of $3$, $5$, and $8$ (implying no decoder). We report in Table~\ref{tab:ablation_depth} the reconstruction error of these models, as well as the fidelity of the consistency mask with the toponym text mask. We observe that shallow encoders have better reconstruction accuracy while the consistency masks of deeper encoders better approximate the text masks. 

Visualizing these masks in~\Cref{fig:PlanIGN_ablation_mask}, we observe that shallow encoders consider  
complex features such as highways and rivers as unmatchable features, while deeper encoders do not. Shallower encoders seem more influenced by the variation in appearance (\eg, rivers being sometimes covered in vegetation or with varying colors) while deeper encoders focus on high-level semantics. We argue that both definitions are equally valid, and that varying the depth of the encoders can provide insights into the nature of the semantic mismatch between datasets.

\vspace{0.25cm}\noindent\textbf{Parameterization.} In~\Cref{tab:ablation:reg}, we analyze the effects of omitting the additional terms $\mathcal{L}_\text{reg}$ and $\mathcal{L}_\text{match}$ from the loss function $\mathcal{L}$ in~\Cref{eq:loss:cycleGAN:gene}. 
We show that while $\mathcal{L}_\text{match}$ generally yields modest improvements across all metrics, $\mathcal{L}_\text{reg}$ is pivotal, particularly for learning the target distribution. 
This outcome aligns with our expectations, as the absence of $\mathcal{L}_\text{reg}$ allows the network to transmit all information, matchable or not, to the $\GYX$ decoder without repercussions, impeding the training of the $\GXY$ encoder.

\vspace{0.25cm}\noindent\textbf{Limitations.} We augment the CycleGAN framework with a module $M$ and two hyper-parameters $\lambda_\text{match}$ and $\lambda_\text{reg}$, thereby adding to the complexity of its training dynamics. Moreover, the concept of unmatchability, integral to our approach, is inherently subjective. Given enough semantic detail, any two distinct datasets could be considered unmatchable. 
As a result, fine-tuning the hyperparameter $\lambda_\text{reg}$ is essential to balance the elimination of hallucinations against the retention of necessary details: increasing its value leads to more conservative masks and improves the visual appearance of the generated images, at the cost of more spurious features.  Visualizing the consistency mask is often a useful form of guidance. 
More details on learning strategies can be found in the Appendix.

\input{tex_table/Google_ablation}

We also acknowledge the potential of recent denoising diffusion models for image-to-image translation tasks~\cite{xia2023diffi2i, li2023bbdm}. 
While our method is not confined to GANs and could be adapted to diffusion models with cycle consistency losses~\cite{cyclediffusion,su2022dual}, unpaired image translation with diffusion models is a nascent field with unique challenges. 
We plan to explore this area in future research.

%% file: tex_table/results_on_google_map.tex
\definecolor{STEGOCOLOR}{RGB}{30, 150, 252}
\definecolor{SRUNITCOLOR}{RGB}{249, 65, 68}
\definecolor{GCGANCOLOR}{RGB}{184, 146, 255}
\definecolor{DRITCOLOR}{RGB}{255, 194, 226}
\definecolor{CYCLEGANCOLOR}{RGB}{244, 140, 6}
\definecolor{CUTCOLOR}{RGB}{144, 190, 109}

\begin{figure*}[!h]
\centering
\hspace{-5mm}
 \resizebox{.95\textwidth}{!}{
    \begin{tabular}{c@{}c@{}c@{}c}
    \begin{tikzpicture}
        \begin{axis}[hide axis, xmin=0, xmax=1, ymin=0, ymax=1, legend style={draw=none, legend cell align=left},
    width=.30\linewidth,
    height=.2\textheight]
        \addlegendimage{STEGOCOLOR, mark=*, very thick}
        \addlegendentry{\bf StegoGAN (ours)}
        \addlegendimage{CYCLEGANCOLOR, mark=x, very thick}
        \addlegendentry{CycleGAN \protect\cite{CycleGAN2017}}
        \addlegendimage{DRITCOLOR, mark=triangle, very thick}
        \addlegendentry{DRIT \protect\cite{Lee2018DRIT}}
        \addlegendimage{GCGANCOLOR, mark=o, very thick}
        \addlegendentry{GcGAN \protect\cite{FuCVPR19-GcGAN}}
        \addlegendimage{CUTCOLOR, mark=+, very thick}
        \addlegendentry{CUT \protect\cite{park2020cut}}        
        \addlegendimage{SRUNITCOLOR, mark=square, very thick}
        \addlegendentry{SRUNIT \protect\cite{jia2021srunit}}
    \end{axis}
    \vspace{1cm}
\end{tikzpicture}
    &
\begin{tikzpicture}
\begin{axis}[
    ylabel style={at={(axis description cs:+0.2,0.5)},anchor=north},
    ylabel={$\leftarrow$ RMSE },
    xmin=0, xmax=70,
    ymin=20, ymax=35,
    xtick={0, 25, 45, 65},
    ytick={20, 35},
    legend pos=north west,
    ymajorgrids=true,
    grid style=dashed,
    width=.30\linewidth,
    height=.2\textheight
]

\addplot[color=STEGOCOLOR, mark=*, very thick] coordinates {
    (0, 22.2)
    (25, 23.0)
    (45, 23.6)
    (65, 22.7)
};

\addplot[color=SRUNITCOLOR, mark=square, very thick] coordinates {
    (0, 23.6)
    (25, 24.9)
    (45, 25.3)
    (65, 28.9)
};

\addplot[color=GCGANCOLOR, mark=o, very thick] coordinates {
    (0, 28.9)
    (25, 30.9)
    (45, 29.7)
    (65, 30.1)
};

\addplot[color=DRITCOLOR, mark=triangle, very thick] coordinates {
    (0, 28.5)
    (25, 27.7)
    (45, 27.6)
    (65, 27.9)
};

\addplot[color=CYCLEGANCOLOR, mark=x, very thick] coordinates {
    (0, 21.6)
    (25, 23.7)
    (45, 23.9)
    (65, 25.5)
};

\addplot[color=CUTCOLOR, mark=+, very thick] coordinates {
    (0, 24.6)
    (25, 24.7)
    (45, 25.4)
    (65, 26.2)
};

\end{axis}
\end{tikzpicture}
         & 
        \begin{tikzpicture}
\begin{axis}[
    ylabel style={at={(axis description cs:+0.2,0.5)},anchor=north},
    ylabel={Acc($\sigma_1$) $\rightarrow$},
    xmin=0, xmax=70,
    ymin=20, ymax=45,
    xtick={0, 25, 45, 65},
    ytick={20, 45},
    legend pos=north west,
    ymajorgrids=true,
    grid style=dashed,
    width=.30\linewidth,
    height=.2\textheight
]

\addplot[color=SRUNITCOLOR, mark=square, very thick] coordinates {
    (0, 38.0)
    (25, 37.7)
    (45, 35.4)
    (65, 35.7)
};

\addplot[color=GCGANCOLOR, mark=o, very thick] coordinates {
    (0, 26.2)
    (25, 20.6)
    (45, 24.6)
    (65, 23.4)
};

\addplot[color=DRITCOLOR, mark=triangle, very thick] coordinates {
    (0, 22.5)
    (25, 24.4)
    (45, 27.6)
    (65, 27.1)
};

\addplot[color=CYCLEGANCOLOR, mark=x, very thick] coordinates {
    (0, 43.6)
    (25, 41.0)
    (45, 42.6)
    (65, 41.0)
};

\addplot[color=CUTCOLOR, mark=+, very thick] coordinates {
    (0, 39.0)
    (25, 39.4)
    (45, 37.0)
    (65, 37.5)
};

\addplot[color=STEGOCOLOR, mark=*, very thick] coordinates {
    (0, 42.2)
    (25, 42.3)
    (45, 41.7)
    (65, 41.7)
};

\end{axis}
\end{tikzpicture}
&
\begin{tikzpicture}
\begin{axis}[
    ylabel style={at={(axis description cs:+0.2,0.5)},anchor=north},
    ylabel={Acc($\sigma_2$) $\rightarrow$},
    xmin=0, xmax=70,
    ymin=40, ymax=70,
    xtick={0, 25, 45, 65},
    ytick={40, 70},
    legend pos=north west,
    ymajorgrids=true,
    grid style=dashed,
    width=.30\linewidth,
    height=.2\textheight
]

\addplot[color=SRUNITCOLOR, mark=square, very thick] coordinates {
    (0, 63.4)
    (25, 62.4)
    (45, 61.4)
    (65, 59.8)
};

\addplot[color=GCGANCOLOR, mark=o, very thick] coordinates {
    (0, 45.4)
    (25, 40.6)
    (45, 44.4)
    (65, 44.0)
};

\addplot[color=DRITCOLOR, mark=triangle, very thick] coordinates {
    (0, 43.8)
    (25, 45.0)
    (45, 47.3)
    (65, 47.2)
};

\addplot[color=CYCLEGANCOLOR, mark=x, very thick] coordinates {
    (0, 68.1)
    (25, 64.8)
    (45, 66.2)
    (65, 65.3)
};

\addplot[color=CUTCOLOR, mark=+, very thick] coordinates {
    (0, 61.6)
    (25, 61.6)
    (45, 59.9)
    (65, 61.0)
};

\addplot[color=STEGOCOLOR, mark=*, very thick] coordinates {
    (0, 67.5)
    (25, 66.3)
    (45, 65.3)
    (65, 67.1)
};

\end{axis}
\end{tikzpicture}
\\
\begin{tikzpicture}
\begin{axis}[
    ylabel style={at={(axis description cs:+0.2,0.5)},anchor=north},
    ylabel={$\leftarrow$ pFPR (\textpertenthousand) },
    xmin=0, xmax=70,
    ymin=0, ymax=20,
    xtick={0, 25, 45, 65},
    ytick={0, 20},
    yticklabels={0,20},
    extra tick style={tick style={draw=none}},
    legend pos=north west,
    ymajorgrids=true,
    grid style=dashed,
    width=.30\linewidth,
    height=.2\textheight
]

\addplot[color=SRUNITCOLOR, mark=square, very thick] coordinates {
    (0, 0)
    (25, 2.1)
    (45, 2.8)
    (65, 38.1)
};

\addplot[color=GCGANCOLOR, mark=o, very thick] coordinates {
    (0, 0)
    (25, 1.2)
    (45, 3.9)
    (65, 7.9)
};

\addplot[color=DRITCOLOR, mark=triangle, very thick] coordinates {
    (0, 0)
    (25, 0)
    (45, 0.3)
    (65, 1.6)
};

\addplot[color=CYCLEGANCOLOR, mark=x, very thick] coordinates {
    (0, 0)
    (25, 0.4)
    (45, 14.1)
    (65, 24.9)
};

\addplot[color=CUTCOLOR, mark=+, very thick] coordinates {
    (0, 0)
    (25, 0.9)
    (45, 3.9)
    (65, 29.8)
};

\addplot[color=STEGOCOLOR, mark=*, very thick] coordinates {
    (0, 0)
    (25, 0)
    (45, 0)
    (65, 0)
};

\end{axis}
\end{tikzpicture}
&
\begin{tikzpicture}
\begin{axis}[
    ylabel style={at={(axis description cs:+0.2,0.5)},anchor=north},
    ylabel={$\leftarrow$ iFPR (\%) },
    xmin=0, xmax=70,
    ymin=0, ymax=20,
    xtick={0, 25, 45, 65},
    ytick={0, 20},
    yticklabels={0,20},
    extra tick style={tick style={draw=none}},
    legend pos=north west,
    ymajorgrids=true,
    grid style=dashed,
    width=.30\linewidth,
    height=.2\textheight
]

\addplot[color=SRUNITCOLOR, mark=square, very thick] coordinates {
    (0, 0.1)
    (25, 3.6)
    (45, 7.9)
    (65, 31.5)
};

\addplot[color=GCGANCOLOR, mark=o, very thick] coordinates {
    (0, 0)
    (25, 18.7)
    (45, 5.9)
    (65, 20.5)
};

\addplot[color=DRITCOLOR, mark=triangle, very thick] coordinates {
    (0, 0)
    (25, 0)
    (45, 0.8)
    (65, 1.9)
};

\addplot[color=CYCLEGANCOLOR, mark=x, very thick] coordinates {
    (0, 0)
    (25, 0.9)
    (45, 13.1)
    (65, 17.1)
};

\addplot[color=CUTCOLOR, mark=+, very thick] coordinates {
    (0, 0)
    (25, 1.4)
    (45, 7.2)
    (65, 20.2)
};

\addplot[color=STEGOCOLOR, mark=*, very thick] coordinates {
    (0, 0)
    (25, 0)
    (45, 0)
    (65, 0)
};

\end{axis}
\end{tikzpicture}
&
\begin{tikzpicture}
\begin{axis}[
    ylabel style={at={(axis description cs:+0.2,0.5)},anchor=north},
    ylabel={$\leftarrow$ FID },
    xmin=0, xmax=70,
    ymin=70, ymax=100,
    xtick={0, 25, 45, 65},
    ytick={70, 100},
    yticklabels={70,100},
    extra tick style={tick style={draw=none}},
    legend pos=north west,
    ymajorgrids=true,
    grid style=dashed,
    width=.30\linewidth,
    height=.2\textheight
]

\addplot[color=SRUNITCOLOR, mark=square, very thick] coordinates {
    (0, 70.6)
    (25, 78.5)
    (45, 81.8)
    (65, 97.7)
};

\addplot[color=GCGANCOLOR, mark=o, very thick] coordinates {
    (0, 85.3)
    (25, 171.0)
    (45, 98.7)
    (65, 123.1)
};

\addplot[color=DRITCOLOR, mark=triangle, very thick] coordinates {
    (0, 204.6)
    (25, 152.8)
    (45, 142.5)
    (65, 143.8)
};

\addplot[color=CYCLEGANCOLOR, mark=x, very thick] coordinates {
    (0, 70.8)
    (25, 71.4)
    (45, 88.1)
    (65, 92.1)
};

\addplot[color=CUTCOLOR, mark=+, very thick] coordinates {
    (0, 66)
    (25, 71.1)
    (45, 78.9)
    (65, 99.7)
};

\addplot[color=STEGOCOLOR, mark=*, very thick] coordinates {
    (0, 70.6)
    (25, 70.1)
    (45, 78.7)
    (65, 77.3)
};

\end{axis}
\end{tikzpicture}
&
\begin{tikzpicture}
\begin{axis}[
    ylabel style={at={(axis description cs:+0.2,0.5)},anchor=north},
    ylabel={$\leftarrow$ KID },
    xmin=0, xmax=70,
    ymin=4, ymax=10,
    xtick={0, 25, 45, 65},
    ytick={4, 10},
    yticklabels={4,10},
    extra tick style={tick style={draw=none}},
    legend pos=north west,
    ymajorgrids=true,
    grid style=dashed,
    width=.30\linewidth,
    height=.2\textheight
]

\addplot[color=SRUNITCOLOR, mark=square, very thick] coordinates {
    (0, 5.8)
    (25, 6.3)
    (45, 6.9)
    (65, 7.9)
};

\addplot[color=GCGANCOLOR, mark=o, very thick] coordinates {
    (0, 6.0)
    (25, 20.3)
    (45, 9.1)
    (65, 11.9)
};

\addplot[color=DRITCOLOR, mark=triangle, very thick] coordinates {
    (0, 21.7)
    (25, 14.0)
    (45, 13.1)
    (65, 13.3)
};

\addplot[color=CYCLEGANCOLOR, mark=x, very thick] coordinates {
    (0, 6.0)
    (25, 6.1)
    (45, 8.6)
    (65, 8.9)
};

\addplot[color=CUTCOLOR, mark=+, very thick] coordinates {
    (0, 4.9)
    (25, 5.4)
    (45, 6.1)
    (65, 8.7)
};

\addplot[color=STEGOCOLOR, mark=*, very thick] coordinates {
    (0, 5.6)
    (25, 5.8)
    (45, 7.1)
    (65, 6.8)
};

\end{axis}
\end{tikzpicture}
\\
\multicolumn{4}{c}{Percentage of images with highways in the target domain.}
    \end{tabular}
    }
    \caption{{\bf Results on GoogleMaps.} We report the performance of several top-performing image translation models for different ratios of unmatchable features in the target domain of the training set. StegoGAN handles higher ratios better than competing methods.}
\label{fig:tex_table/google_results_graph}
\end{figure*}
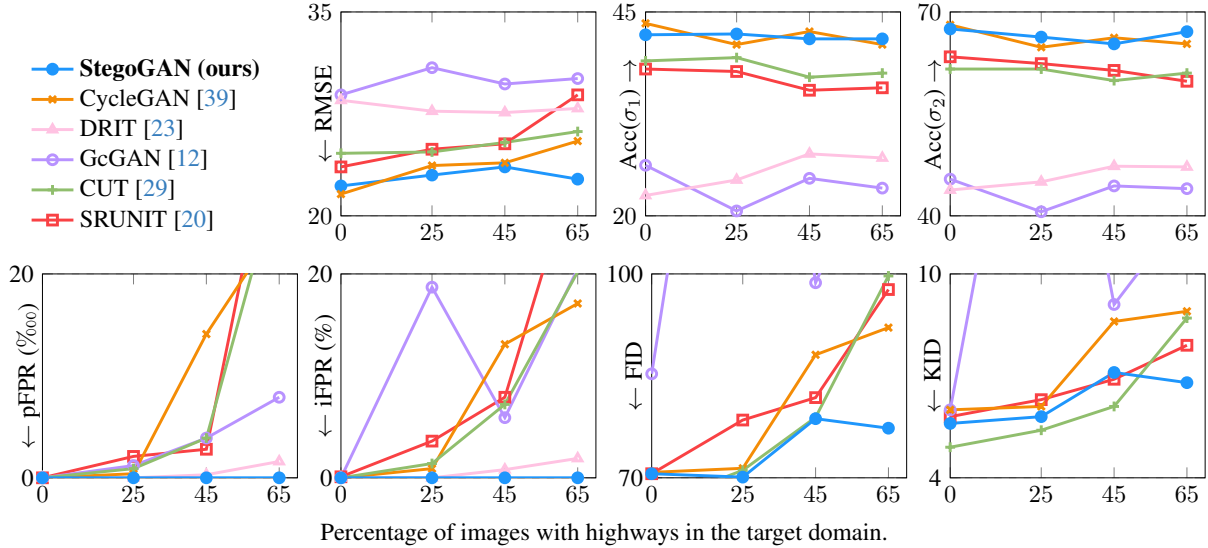

%% file: tex_table/PlanIGN_results.tex
\begin{table}[t]
\centering
\caption{\textbf{Quantitative Comparison on PlanIGN}. Our model shows a remarkably better performance than other existing models.}
\vspace{-0.5em}
\resizebox{\linewidth}{!}{
\begin{tabular}{cccccc}
\toprule
Method &  RMSE$\downarrow$ &  Acc($\sigma_1$)$\uparrow$ &  Acc($\sigma_2$)$\uparrow$ &    FID$\downarrow$ &  KID$\downarrow$ \\
\midrule
CUT \cite{park2020cut} &  30.5 &  46.7 &  55.8 &  68.4 &  2.8 \\
\rowcolor{gray!25}
CycleGAN \cite{CycleGAN2017} &  27.0 &  15.1 &  57.6 &  97.5 &  6.6 \\
DRIT \cite{Lee2018DRIT} &  34.8 &  33.6 &  36.9 &  76.4 &  3.8 \\
\rowcolor{gray!25}
GcGAN \cite{FuCVPR19-GcGAN} &  32.7 &  54.5 &  56.9 & 110.8 &  8.2 \\
SRUNIT \cite{jia2021srunit} &  32.3 &  48.8 &  52.8 &  60.2 &  \textbf{2.2} \\
\rowcolor{gray!25}
\bf StegoGAN (ours) &  \textbf{22.5} &  \textbf{66.1} &  \textbf{74.8} &  \textbf{58.4} &  2.4 \\
\bottomrule
\end{tabular}
}
\label{tab:planIGN_results}
\end{table}

%% file: tex_table/Brats_result_0.6.tex
\begin{table}
\centering
\caption{\textbf{Quantitative Comparison on Brats MRI Flair $\rightarrow$ T1.} Our model outperforms competing method in terms of both reconstruction accuracy and consistency. }
\vspace{-0.5em}
\resizebox{\linewidth}{!}{
\begin{tabular}{cccccc}
\toprule
Method &  RMSE$\downarrow$ &  pFPR(\textpertenthousand)$\downarrow$ & iFPR$\downarrow$ &   FID$\downarrow$ &  KID$\downarrow$ \\
\midrule
CUT \cite{park2020cut} &  39.8 &     17.0 &  23.0 & 103.9 &  8.8 \\
\rowcolor{gray!25}
CycleGAN \cite{CycleGAN2017} &  39.9 &     21.9 &  22.7 &  89.8 &  7.7 \\
DRIT \cite{Lee2018DRIT} &  53.0 &     18.5 &  41.2 & 123.4 & 11.6 \\
\rowcolor{gray!25}
GcGAN \cite{FuCVPR19-GcGAN} &  41.7 &     24.7 &  22.4 &  61.9 &  3.5 \\
SRUNIT \cite{jia2021srunit} &  42.5 &     15.1 &  21.8 &  58.9 &  2.9 \\
\rowcolor{gray!25}
\bf StegoGAN (ours) &   \textbf{36.3} & \textbf{1.1} &   \textbf{4.2} &  \textbf{58.5} &  \textbf{2.4} \\
\bottomrule
\end{tabular}
}
\label{tab:quantitativetumor}
\end{table} 

%% file: figures/unmatchability_masks.tex
\begin{figure}[t]
  \centering
  \setlength\tabcolsep{1pt}
  \begin{tabular}{cccc}
    Target & Source & Mask \\
    \includegraphics[width=0.14\textwidth]{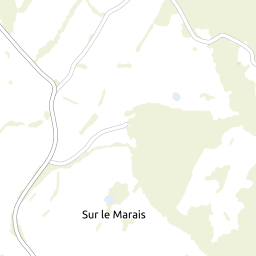}&
    \includegraphics[width=0.14\textwidth]{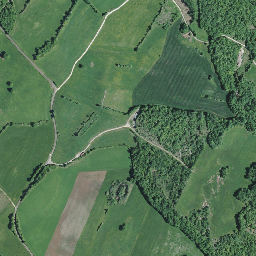}&
    \includegraphics[width=0.14\textwidth]{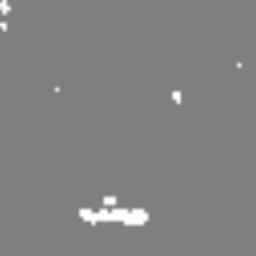}
    \\
    \includegraphics[width=0.14\textwidth]{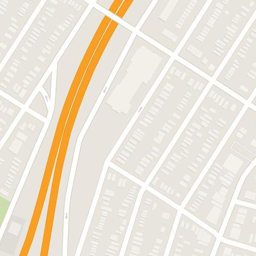}&
    \includegraphics[width=0.14\textwidth]{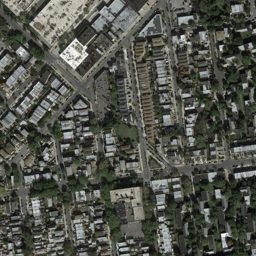}&
    \includegraphics[width=0.14\textwidth]{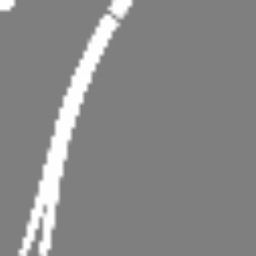}\\
    \vspace{-2em}
    \\
    \includegraphics[angle=-90, width=0.14\textwidth]{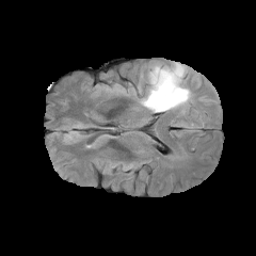}&
    \includegraphics[angle=-90, width=0.14\textwidth]{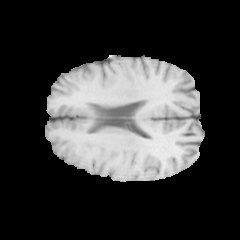}&
    \includegraphics[angle=-90, width=0.14\textwidth]{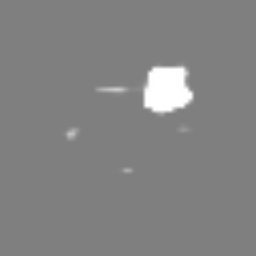}
  \end{tabular}
  \caption{ {\bf Unmatchability Masks.} The unmatchability masks predicted in the backward cycle follow the instances of unmatchable features in the target domain: toponyms, highways, and tumors.}
  \label{fig:visual_mask}
\end{figure}

%% file: tex_table/PlanIGN_ablation_depth.tex
\begin{table}[tb]
\centering
\caption{\textbf{Ablation Study on Encoder Depth.}  We evaluate the impact of changing the depth of the encoder on the reconstruction fidelity and the quality of unmatchability masks. Depth= -1 or 8 means no encoder or no decoder, respectively.
}
\vspace{-0.5em}
\scriptsize
\begin{tabularx}{\linewidth}{lXXXcXccXc}
\toprule
\multirow{3}{*}{Depth} & \multicolumn{3}{c}{Mask} & \multicolumn{5}{c}{Prediction}  \\ \cmidrule(r{4pt}){2-4} \cmidrule(r{4pt}){5-9} 
   &  mIOU$\uparrow$ &  Prec.$\uparrow$ &  recall$\uparrow$&  RMSE$\downarrow$ &  A($\sigma_1$)$\uparrow$&  A($\sigma_2$)$\uparrow$ &  FID$\downarrow$ &  KID$\downarrow$\\
\midrule
     -1 &        26.6 &             27.1 &         \textbf{81.2} &  \textbf{22.4} &  64.3 &  74.2 & 58.8 &  2.5 \\
    \rowcolor{gray!25}
     1 &        25.2 &             25.8 &          \textbf{81.2} &  22.5 &  \textbf{66.1} &  \textbf{74.8} & \textbf{58.4} &  \textbf{2.4} \\
     3 &        27.1 &             27.9 &          80.4 &  22.8 &  61.6 &  73.4 & 62.9 &  3.0 \\
     \rowcolor{gray!25}
     5 &        30.8 &             33.4 &          69.5 &  24.2 &  52.7 &  70.6 & 62.3 &  2.6 \\
     8 &        \textbf{47.3} &     \textbf{60.1}  &      60.0 &  24.5 &  53.5 &  70.7 & 62.7 &  2.7\\
\bottomrule
\end{tabularx}
\label{tab:ablation_depth}
\end{table}

%% file: figures/Ablation_mask.tex
\begin{figure*}[ht]
  \centering
  \setlength\tabcolsep{1pt}
  \resizebox{\linewidth}{!}{
  \begin{tabular}{ccccccc}
    Source & Target & Depth = -1 & Depth = 1 & Depth = 3 & Depth = 5 & Depth = 8  \\
    \includegraphics[width=0.14\textwidth]{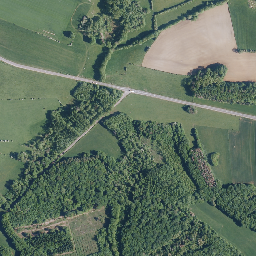}&
    \includegraphics[width=0.14\textwidth]{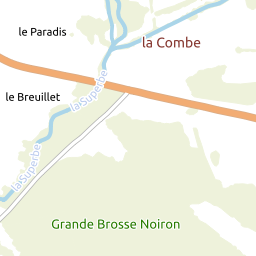}&
    \includegraphics[width=0.14\textwidth]{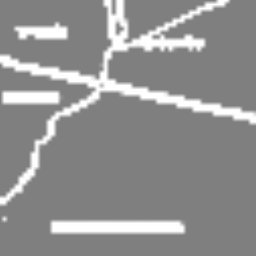}&
    \includegraphics[width=0.14\textwidth]{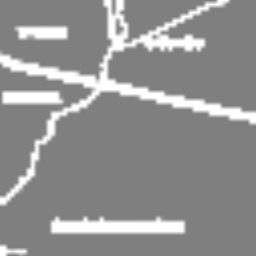}&
    \includegraphics[width=0.14\textwidth]{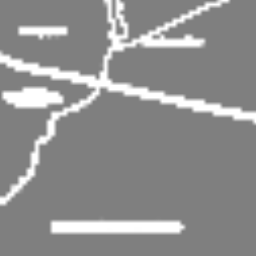}&
    \includegraphics[width=0.14\textwidth]{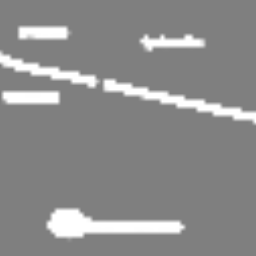}&
    \includegraphics[width=0.14\textwidth]{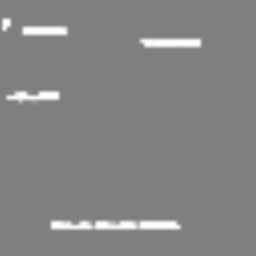}\\
  \end{tabular}
  }
  \caption{{\bf Impact of Encoder Depth.} %
  We visualize the unmatchability mask for encoders of different depths for the PlanIGN dataset. Shallower encoders consider more features as unmatchable.  }
  \label{fig:PlanIGN_ablation_mask}
\vspace{0.1cm}
\end{figure*}

%% file: tex_table/Google_ablation.tex
\begin{table}[t]
\centering
\caption{{\bf Impact of Additional Loss Terms.}
We evaluate on the GoogleMap dataset the effect of removing our proposed losses. $\mathcal{L}_{\text{match}}$ has a small impact while $\mathcal{L}_{\text{reg}}$ is pivotal.}
\label{tab:ablation:reg}
\resizebox{\linewidth}{!}{
\begin{tabular}{ccccccccc}
\toprule
\multicolumn{2}{c}{Settings} &  \multirow{2}{*}{RMSE$\downarrow$} &  \multirow{2}{*}{Acc($\sigma_1$)$\uparrow$} &  \multirow{2}{*}{Acc($\sigma_2$)$\uparrow$} &   \multirow{2}{*}{FID$\downarrow$} &  \multirow{2}{*}{KID$\downarrow$} \\ \cline{1-2}
$\mathcal{L}_{\text{reg}}$ & $\mathcal{L}_{\text{match}}$ & & & & &         \\
\midrule
\cmark & \cmark & 22.7 &  41.7 &   67.1 &  77.3 &  6.8 \\
\rowcolor{gray!25}
\cmark & \xmark & 24.1 &  41.5 &   64.7 &  88.0 &  7.5 \\
\xmark & \cmark & 26.7 &  26.7 &   58.9 &  271.1 & 26.6 \\
\rowcolor{gray!25}
\xmark & \xmark & 25.0 &  25.0 &   61.0 &  303.6 & 33.4 \\
\bottomrule
\end{tabular}
}
\end{table}

%% file: sec/5_conclusion.tex
\section{Conclusions}
\label{sec:conclusions}

We have introduced StegoGAN, a model built upon the CycleGAN framework, which leverages the mechanism of steganography to address the challenges of non-bijective image-to-image translation. Our model demonstrates an improved capability to handle divergent distributions between domains,  as evidenced by its performance across various datasets, including aerial imagery, topographic maps, and MRI scans. We hope that our work will inspire further research in the little-studied area of non-bijective image translation. We find this research direction inportant to ensure image translation models are transferable and applicable in real-world scenarios, where datasets rarely conform to the level of curation typically found in research benchmarks.

%% file: sec/X_suppl.tex
\clearpage
\setcounter{page}{1}
\setcounter{section}{0}
\setcounter{table}{0}
\setcounter{figure}{0}

\renewcommand{\thefigure}{A-\arabic{figure}}
\renewcommand{\thetable}{A-\arabic{table}}
\maketitlesupplementary
\vspace{1em}

\noindent
In this appendix, we first detail the construction of the three datasets used in the experiments (Section \ref{sup:sec:datasets}), then provide additional implementation details (Section \ref{sup:implem}), ablation experiments (Section \ref{sup:ablation}), and qualitative results (Section \ref{sup:quali}).

\section{Details on Dataset Construction}

\label{sup:sec:datasets}

\subsection{GoogleMaps}
\label{sup:sec:datasets:google}
While this paper focuses on non-bijective translation, the official GoogleMaps dataset was introduced in CycleGAN \cite{CycleGAN2017} for bijective translation between serial photos and maps. Consequently, we propose a protocol to create controllable non-bijectivity in this dataset.

We select the ``highway'' class for its prevalence and its distinctiveness on maps: they are always represented by the same orange hue. This allows us to easily detect highways on maps by thresholding in color space: a pixel is a highway if all color channels are closer than $20$ units from $(240,160,30)$. In total, $356$ image/map pairs of the train set of the GoogleMaps contain highways, and $740$ do not. 

For the training set, the source domain is always defined as $548$ aerial images that do not contain highways. We define different versions of the target domain for the test set by fixing the ratio of maps that contains highways, from $0\%$ to $60\%$, for a fixed total of $548$ images.
The test set is composed of $899$ pairs of aligned aerial photos and maps \emph{that do not contain the highways class} from the test set of the GoogleMaps dataset.

\subsection{Brats MRI}
\label{sup:sec:datasets:BRATS}
We adapt the protocol of Cohen \etal~\cite{cohen2018distribution} from the Brats2013 datasets~\cite{menze2014brats13} to the more recent, larger, and more diverse Brats2018 dataset \cite{bakas2017brats17}.
We consider two MRI modalities: native (T1) and  Fluid Attenuated Inversion Recovery (FLAIR).
We selected transverse slices from the $60^\circ$ to $100^\circ$ range in the caudocranial direction~\cite{andermatt2019pathology} for both modalities of scans. 

We label each scan as tumorous if more than $1\%$ of its pixels are labelled as such, and as healthy if it contains no tumor pixels. 
We only use high-grade gliomas (HGG) instead of low-grade gliomas (LGG) as the are more easily observable~\cite{Lefkovitz2022LGGHGG}. In total, we obtain 5035 pathological pairs and 1135 healthy pairs.
The train set is composed of a source domain of $800$ T1 images of healthy brains, while the target domain set is composed of FLAIR scans of which $480$  (60\%) are tumorous and $320$ healthy.
The test set is composed of $335$ aligned scans of healthy brains in both modalities.

\begin{figure}[t]
\includegraphics[width=\linewidth]{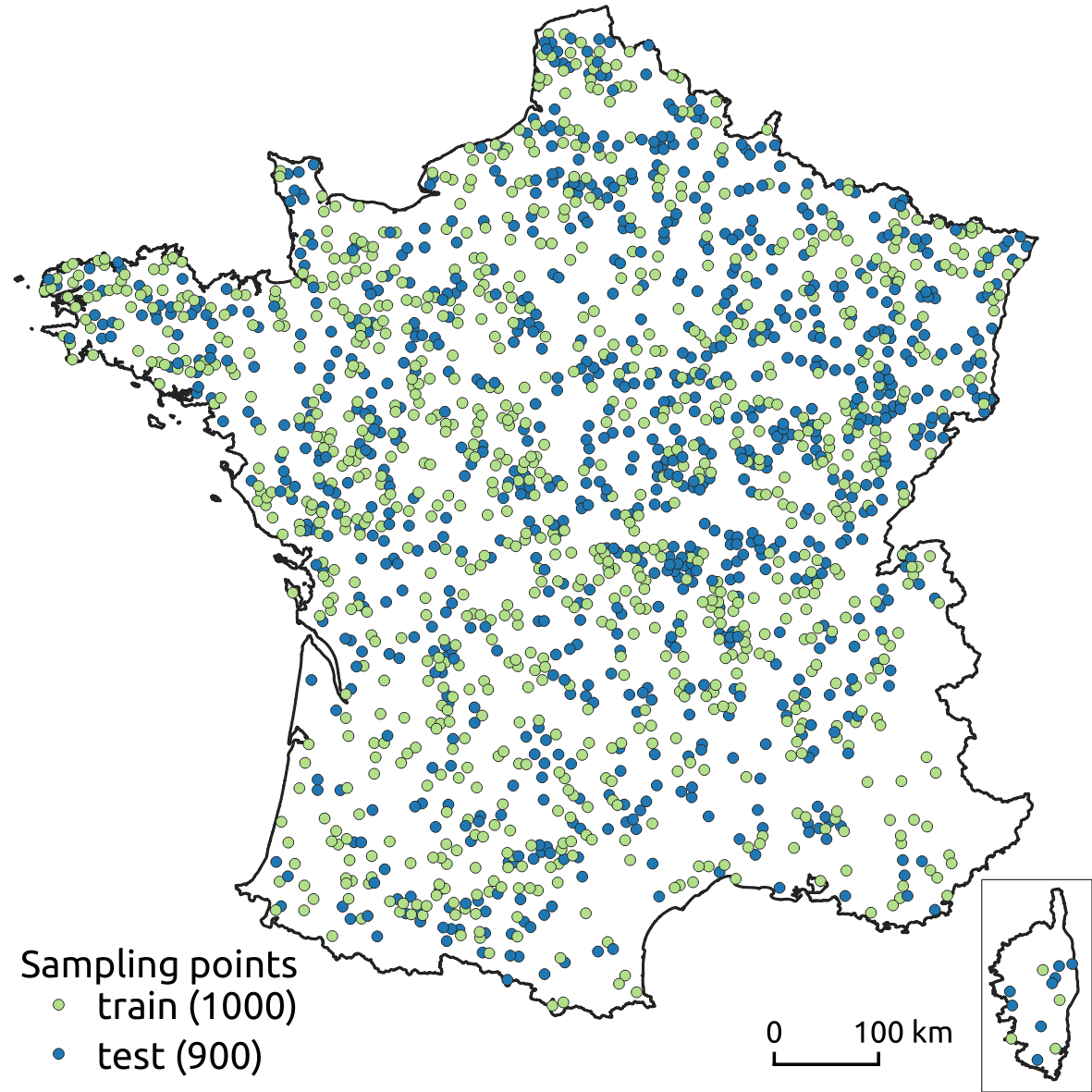}
\caption{{\bf Spatial Distribution of Samples in PlanIGN.}}
\label{fig:sampling_points}
\vspace{2em}
\end{figure}

\subsection{PlanIGN}
\label{sup:sec:datasets:IGN}

We construct the PlanIGN dataset from two open-access sources available on the \href{https://www.geoportail.gouv.fr/}{French governmental geoportal}: aerial orthophotos and Plan IGN cartographic product, both projected in RGF93-Lambert-93. As the maps are derived directly from the orthophotos, we ensure the precise spatial alignment between both modalities.

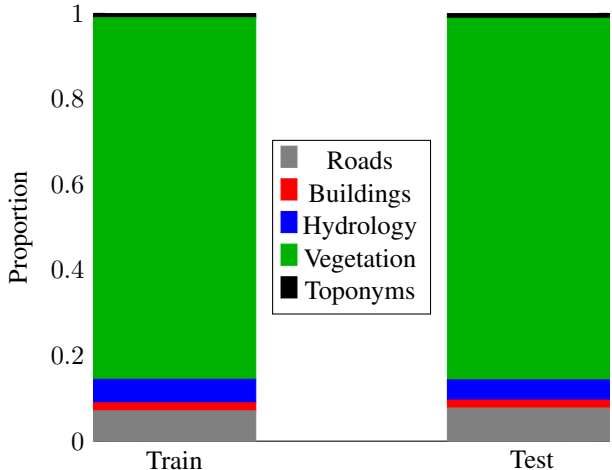
\begin{figure}[t]
\begin{tikzpicture}
\begin{axis}[
    ybar stacked,
    bar width=90pt,
    ylabel={Proportion},
    symbolic x coords={Train, Test},
    xtick={Train, Test},
    xticklabels={\hspace{1cm}Train, \hspace{-1cm} Test},
    xtick align=inside,
    x tick label style={anchor=north ,align=center},
    legend style={at={(0.5,0.5)},
    anchor=center,legend columns=1},
    ylabel near ticks,
    xlabel near ticks,
    ymin=0,ymax=1,
    axis x line*=bottom, %
]

\addplot+[ybar, fill=gray, draw=gray] plot coordinates {(Train, 0.070811) (Test, 0.077409)};
\addplot+[ybar, fill=red, draw=red] plot coordinates {(Train, 0.018738) (Test, 0.018446)};
\addplot+[ybar, fill=blue, draw=blue] plot coordinates {(Train, 0.054925) (Test, 0.047387)};
\addplot+[ybar, fill=green!70!black,draw=green!70!black] plot coordinates {(Train, 0.844827) (Test, 0.844959)};
\addplot+[ybar, fill=black, draw=black] plot coordinates {(Train, 0.010700) (Test, 0.011799)};

\legend{Roads, Buildings, Hydrology, Vegetation, Toponyms}

\end{axis}
\end{tikzpicture}
\caption{ {\bf Semantic Distribution in PlanIGN.}} 
\label{fig:semantic_classes_distribution}
\vspace{1em}
\end{figure}

\paragraph{Sampling.} We consider aligned image/map pairs of resolution $256\times256$ at a scale of 1:12500 and a graphics resolution of 96 dpi, corresponding to a ground sampling distance of $3.3$m per pixel. 
We randomly select samples across the French territory with a $3$km buffer between images. We removed images that were blurry, with significant radiometric aberrations, over sensitive areas, or for which the roads were significantly occluded. In total, we sample $1900$ such pairs, whose spatial distribution is shown in Figure~\ref{fig:sampling_points}, and whose semantic distribution is given in Figure~\ref{fig:semantic_classes_distribution}.

\paragraph{Processing.} We apply the following processing to the maps to make them easier to translate:

\begin{itemize}
    \item We remove underground objects and small paths.
    \item We homogenize the palette: we use the same color for all roads except highways, and for buildings and hydrology.
    \item We optionally add the toponyms to the maps. When we do, we also compute a toponym mask by applying a 4-pixel dilation to the binary difference between the map with and without toponyms.
\end{itemize}
\vspace{-2em}

\paragraph{Dataset.} The training set is composed of $1000$ orthophotos for the source domain and $1000$ maps with toponyms for the target domain. The test set is composed of $900$ aligned pairs of orthophotos and maps without toponyms.

\section{Additional Implementation Details}
\label{sup:implem}
We follow the same architecture and hyperparameters as CycleGAN, including the ResNet-based generator~\cite{he2016res} with $9$ residual blocks, PatchGAN discriminator~\cite{isola2017pix}, and weights in th loss.  We train our model for $200$ epochs with a learning rate of $0.002$ and the ADAM optimizer~\cite{kingma2014adam}. 

The hyperparameters for each dataset are given in Table~\ref{tab:hyperparameters}. Most methods use similar hyperparameters with two exceptions:
\begin{itemize}
    \item Due to the high heterogeneity and noisiness of scans across different MRI machines, we use a larger batch size of $12$.
    \item We observed better unmatchability masks with shallower encoders for PlanIGN, whereas it was the contrary for other datasets. 
    Section 4.4 has pointed out that shallower encoder seems more influenced by the variation in appearance. Despite that class like hydrology exits both in the source and target domain, variations in colors or occlusions by vegetation occurring often in the aerial images challenge the model to establish correct correspondences. This should also be empirically regarded as mismatch, which shallower encoder performs better to capture.
\end{itemize}

\input{tex_table/dataset_parameters}

\pagebreak
\section{Additional Ablation Study}
\label{sup:ablation}
\input{tex_table/Google_ablation_reg}
The hyperparameter $\lambda_\text{reg}$ is crucial as it enforces the sparsity of the unmatchability masks. We report its impact in Table~\ref{tab:ablation_reg}.
Too small values may lead to a too-liberal use of the unmatchability masks, resulting in a loss of details in the clean generation. Too large values will prevent our model from using the unmatchability masks altogether.
\section{Additional Qualitative Results}
\label{sup:quali}
We provide additional results for \textbf{GoogleMaps} in Figure~\ref{fig:Google_qualitative_additional}, \textbf{PlanIGN} in Figure~\ref{fig:PlanIGN_qualitative_additional} and \textbf{Brats MRI} in Figure~\ref{fig:brats_qualitative_additional}.

\begin{figure*}
\newcolumntype{C}{>{\centering\arraybackslash} m{.12\textwidth} }
     \begin{tabular}{c@{\;}c@{\;}c@{\;}c@{\;\,}c@{\;}c@{\;}c@{\;}c}  
     \includegraphics[width=.12\textwidth,height=.09\textheight]{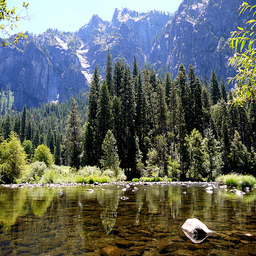}
    & \includegraphics[width=.12\textwidth,height=.09\textheight]{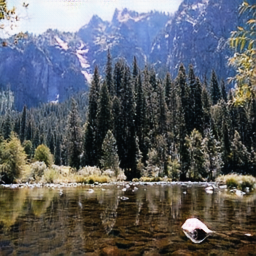}
    & \includegraphics[width=.12\textwidth,height=.09\textheight]{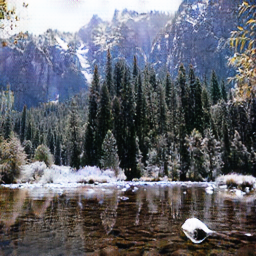}
    & \includegraphics[width=.12\textwidth,height=.09\textheight]{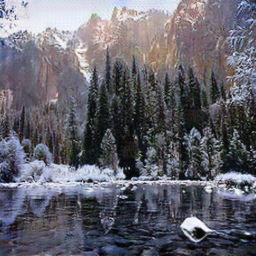} 
    &
    \includegraphics[width=.12\textwidth,height=.09\textheight]{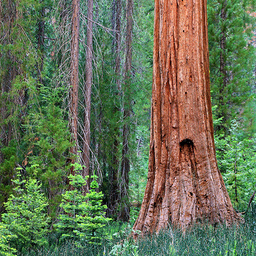}
    & \includegraphics[width=.12\textwidth,height=.09\textheight]{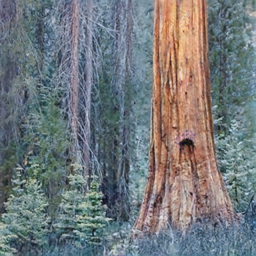}
    & \includegraphics[width=.12\textwidth,height=.09\textheight]{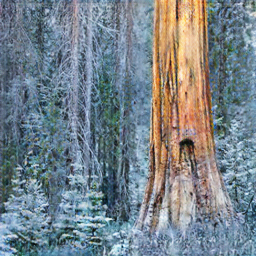}
    & \includegraphics[width=.12\textwidth,height=.09\textheight]{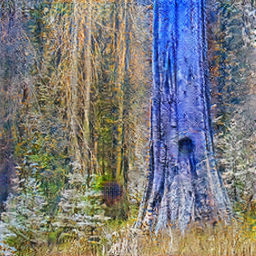} 
    \\
    \includegraphics[width=.12\textwidth,height=.09\textheight]{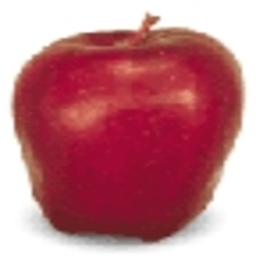}
    & \includegraphics[width=.12\textwidth,height=.09\textheight]{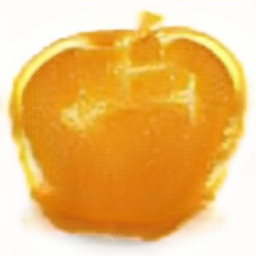}
    & \includegraphics[width=.12\textwidth,height=.09\textheight]{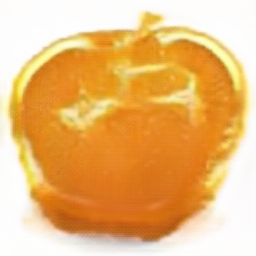}
    & \includegraphics[width=.12\textwidth,height=.09\textheight]{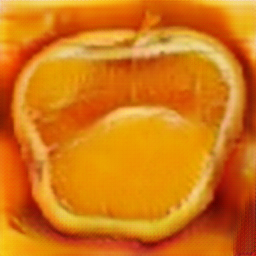} 
    &
    \includegraphics[width=.12\textwidth,height=.09\textheight]{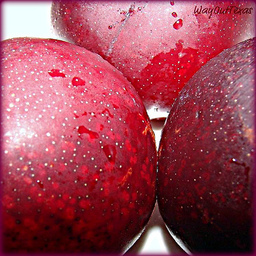}
    & \includegraphics[width=.12\textwidth,height=.09\textheight]{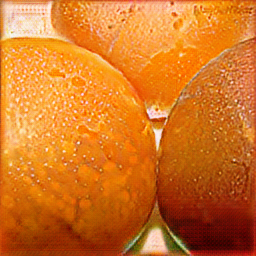}
    & \includegraphics[width=.12\textwidth,height=.09\textheight]{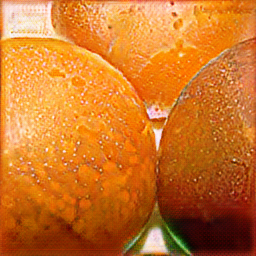}
    & \includegraphics[width=.12\textwidth,height=.09\textheight]{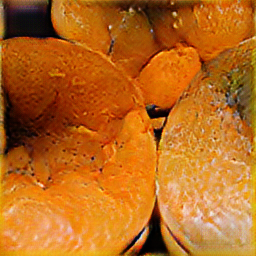} \\
    \cline{2-3}
    \cline{6-7}
    Input & \multicolumn{2}{c}{$y^{\text{clean}}_{\text{gen}}$ \;\; StegoGAN \;\; $y_{\text{gen}}$}  & CycleGAN \cite{CycleGAN2017}
    &
    Input & \multicolumn{2}{c}{$y^{\text{clean}}_{\text{gen}}$ \;\; StegoGAN \;\; $y_{\text{gen}}$}  & CycleGAN \cite{CycleGAN2017}
\end{tabular}
      \caption{{\bf Natural Images Translation.} We apply our model to the {\bf Summer $\mapsto$ Winter Yosemite} (top row) and  {\bf Apple $\mapsto$ Orange} datasets (bottom row).}
    \label{fig:natural}
\end{figure*}

\vspace{1.2mm}\noindent\textbf{Application to Natural Images}. 
We apply our method to natural image datasets; see Fig.~\ref{fig:natural}. StegoGAN performs well in this setting and generates faithful yet realistic images. Compared to CycleGAN, the clean translation $y^{\text{clean}}_{\text{gen}}$ produces fewer unmatchable features like snow or color shifts (Summer $\mapsto$ Winter example), or internal structures of fruits (Apple $\mapsto$ Orange example). 
However, since such features often contribute to the realism and visual appeal of the translated images. Therefore, our method is better suited for domains that value reliability over aesthetics, such as medical images or cartography.

\input{figures/Google_qualitative_additional}

\input{figures/PlanIGN_qualitative_additional}
\input{figures/Brats_qualitative_additional}

%% file: tex_table/dataset_parameters.tex
\begin{table}[]
\caption{{\bf Hyperparameters.} We report the value for different parameters across the datasets used in the experiments.}
\resizebox{\linewidth}{!}{
\begin{tabular}{ccccc}
\toprule
\textbf{Dataset} & $\lambda_\text{reg}$ & Encoder\_Depth & Batch\_Size & $\lambda_\text{match}$ \\
\midrule
\textbf{GoogleMaps}     &   0.3       &      8          &       1     &        1       \\
\textbf{PlanIGN}        &   0.25      &      1          &       1     &        1       \\
\textbf{Brats}          &   0.3       &      8          &       12    &        1       \\
\bottomrule
\end{tabular}
}
\vspace{1em}
\label{tab:hyperparameters}
\end{table}

%% file: tex_table/Google_ablation_reg.tex
\begin{table}
\centering
\caption{{\bf Impact of $\lambda_\text{reg}$.}
We report the performance on the \textbf{GoogleMaps} dataset of our method for different values of regularization strength. Values between $0.3$ and $0.5$ give good results, while the performance rapidly decreases above $0.6$.}
\label{tab:ablation_reg}
\resizebox{\linewidth}{!}{
\begin{tabular}{cccccccc}
\toprule
 $\mathcal{L}_\text{reg}$ &  RMSE$\downarrow$ &  Acc($\sigma_1$)$\uparrow$ &  Acc($\sigma_2$)$\uparrow$ &  pFPR(\textpertenthousand)$\downarrow$  &  iFPR$\downarrow$  &   FID$\downarrow$ &  KID$\downarrow$ \\
\midrule
 0.1 &  26.1 &  11.0 &   60.4 &      0.1 &   7.7 & 254.4 & 25.8 \\
 0.2 &  23.3 &  36.5 &   64.0 &   \textbf{0.0} &   \textbf{0.0} & 141.2 & 17.4 \\
 0.3 &  \textbf{22.7} &  41.7 &   \textbf{67.1} &      \textbf{0.0} &   \textbf{0.0} &  \textbf{77.3} &  \textbf{6.8}\\
 0.4 &  23.5 &  \textbf{41.8} &   65.7 &      \textbf{0.0} &   0.2 &  84.8 &  8.3 \\
 0.5 &  24.4 &  39.5 &   64.6 &      \textbf{0.0} &   1.1 &  86.1 &  8.1 \\
 0.6 &  25.2 &  40.0 &   64.8 &     19.1 &  16.8 & 103.5 & 11.3 \\
 0.7 &  25.7 &  36.8 &   62.7 &     18.8 &  22.2 & 110.1 & 11.3 \\
\bottomrule
\end{tabular}
}
\vspace{2em}
\end{table}

%% file: figures/Google_qualitative_additional.tex
\begin{figure*}[tb]
  \centering
  \setlength\tabcolsep{1pt}
  \begin{tabularx}{\linewidth}{cccccccc}
    Source & Ground Truth & \bf StegoGAN & CycleGAN \cite{CycleGAN2017} & DRIT \cite{Lee2018DRIT} & GcGAN \cite{FuCVPR19-GcGAN} & CUT \cite{park2020cut} & SRUNIT \cite{jia2021srunit}\\
    \includegraphics[width=0.12\linewidth]{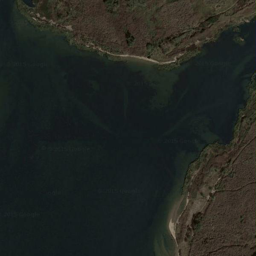}&
    \includegraphics[width=0.12\linewidth]{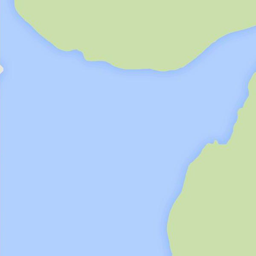}&
    \includegraphics[width=0.12\linewidth]{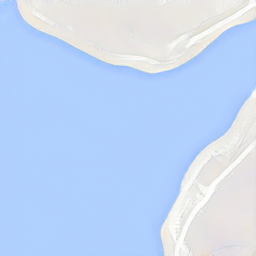}&
    \includegraphics[width=0.12\linewidth]{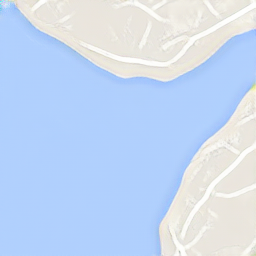}&
    \includegraphics[width=0.12\linewidth]{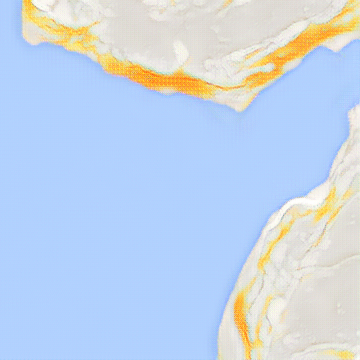}&
    \includegraphics[width=0.12\linewidth]{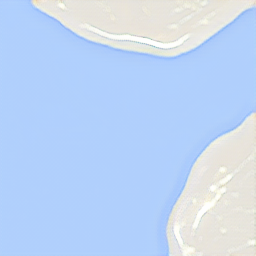}&
    \includegraphics[width=0.12\linewidth]{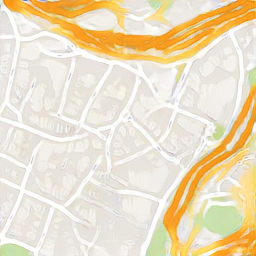} &
    \includegraphics[width=0.12\linewidth]{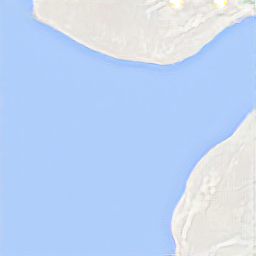}\\
    \includegraphics[width=0.12\linewidth]{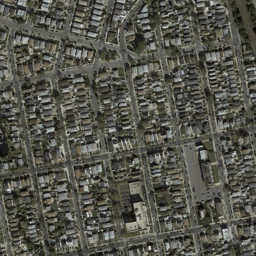}&
    \includegraphics[width=0.12\linewidth]{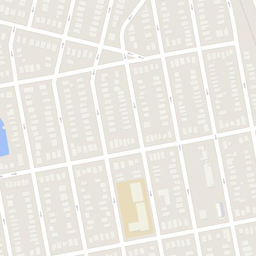}&
    \includegraphics[width=0.12\linewidth]{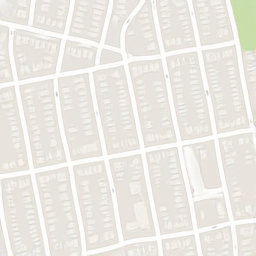}&
    \includegraphics[width=0.12\linewidth]{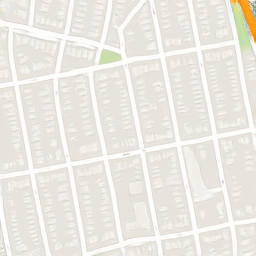}&
    \includegraphics[width=0.12\linewidth]{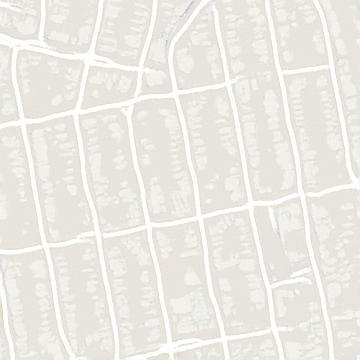}&
    \includegraphics[width=0.12\linewidth]{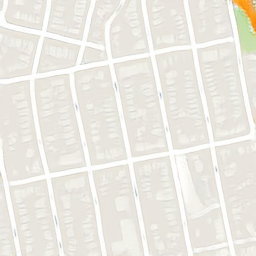}&
    \includegraphics[width=0.12\linewidth]{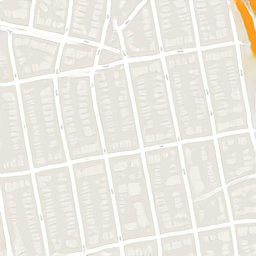} &
    \includegraphics[width=0.12\linewidth]{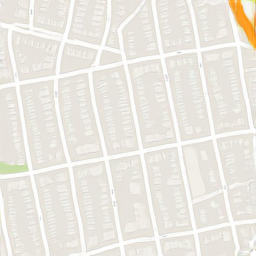}\\
    \includegraphics[width=0.12\linewidth]{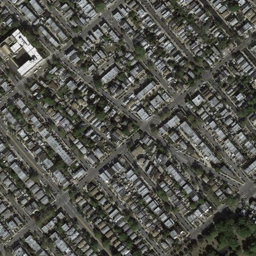}&
    \includegraphics[width=0.12\linewidth]{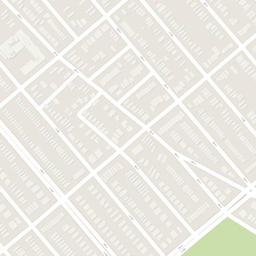}&
    \includegraphics[width=0.12\linewidth]{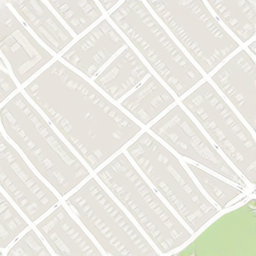}&
    \includegraphics[width=0.12\linewidth]{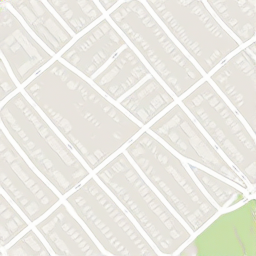}&
    \includegraphics[width=0.12\linewidth]{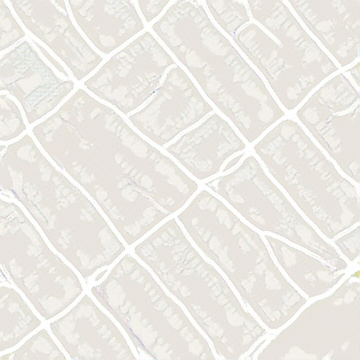}&
    \includegraphics[width=0.12\linewidth]{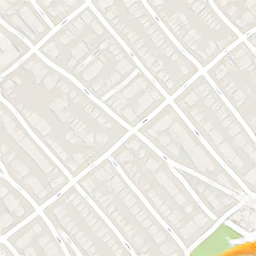}&
    \includegraphics[width=0.12\linewidth]{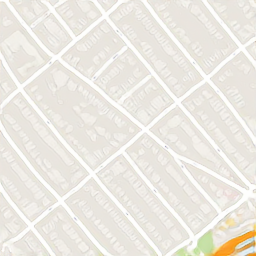} &
    \includegraphics[width=0.12\linewidth]{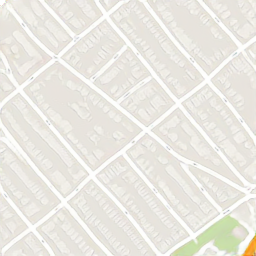}\\
  \end{tabularx}
  \caption{Additional qualitative comparison on \textbf{Google Photo$\rightarrow$Map}.}
  \label{fig:Google_qualitative_additional}
\end{figure*}

%% file: figures/PlanIGN_qualitative_additional.tex
\begin{figure*}[tb]
  \centering
  \setlength\tabcolsep{1pt}
  \begin{tabularx}{\linewidth}{cccccccc}
    Source & Ground Truth & \bf StegoGAN & CycleGAN \cite{CycleGAN2017} & DRIT \cite{Lee2018DRIT} & GcGAN \cite{FuCVPR19-GcGAN} & CUT \cite{park2020cut} & SRUNIT \cite{jia2021srunit}\\
    \includegraphics[width=0.12\linewidth]{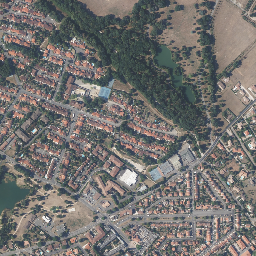}&
    \includegraphics[width=0.12\linewidth]{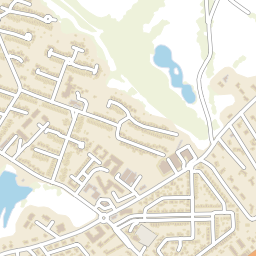}&
    \includegraphics[width=0.12\linewidth]{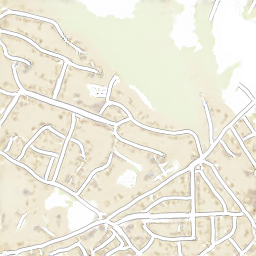}&
    \includegraphics[width=0.12\linewidth]{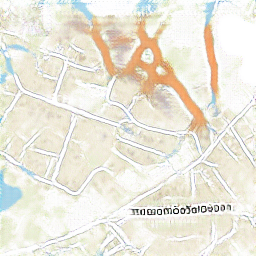}&
    \includegraphics[width=0.12\linewidth]{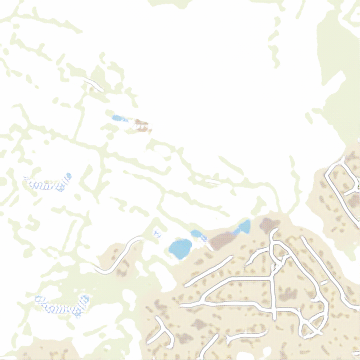}&
    \includegraphics[width=0.12\linewidth]{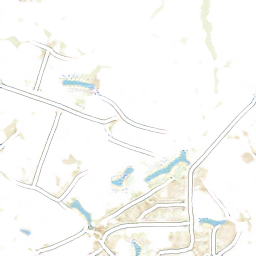}&
    \includegraphics[width=0.12\linewidth]{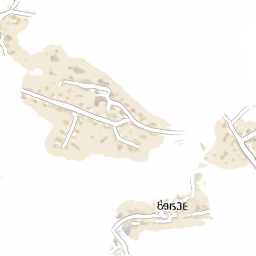}&
    \includegraphics[width=0.12\linewidth]{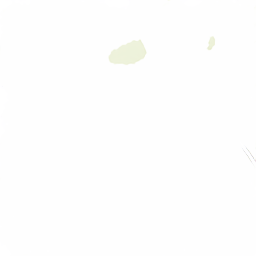}\\
    \includegraphics[width=0.12\linewidth]{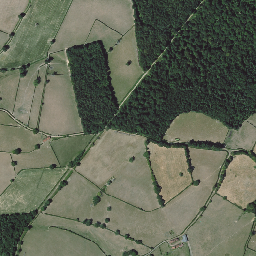}&
    \includegraphics[width=0.12\linewidth]{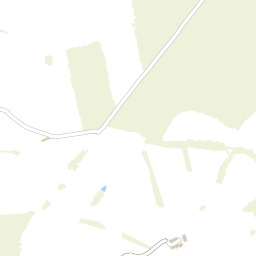}&
    \includegraphics[width=0.12\linewidth]{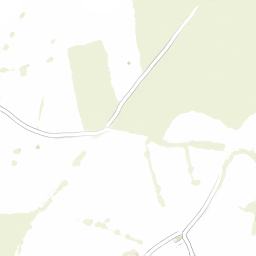}&
    \includegraphics[width=0.12\linewidth]{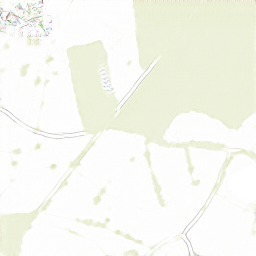}&
    \includegraphics[width=0.12\linewidth]{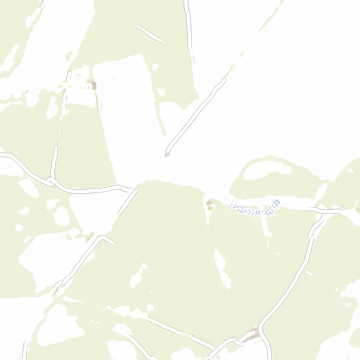}&
    \includegraphics[width=0.12\linewidth]{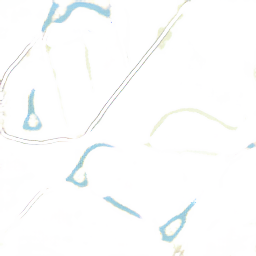}&
    \includegraphics[width=0.12\linewidth]{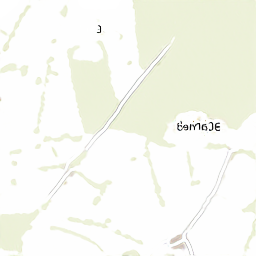}&
    \includegraphics[width=0.12\linewidth]{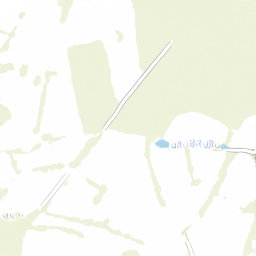}\\
    \includegraphics[width=0.12\linewidth]{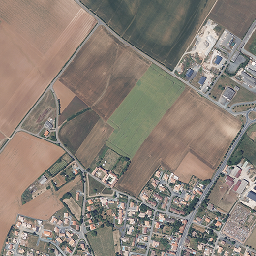}&
    \includegraphics[width=0.12\linewidth]{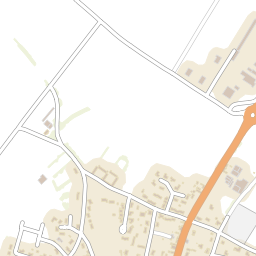}&
    \includegraphics[width=0.12\linewidth]{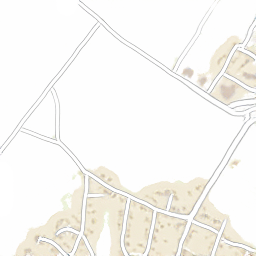}&
    \includegraphics[width=0.12\linewidth]{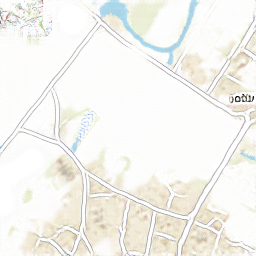}&
    \includegraphics[width=0.12\linewidth]{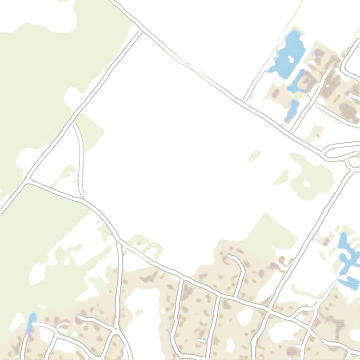}&
    \includegraphics[width=0.12\linewidth]{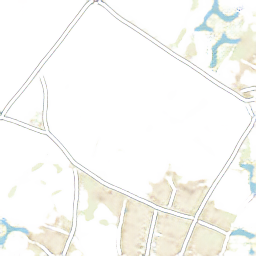}&
    \includegraphics[width=0.12\linewidth]{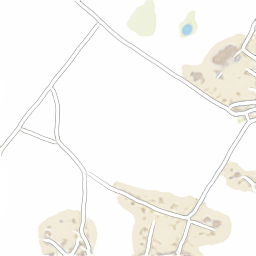}&
    \includegraphics[width=0.12\linewidth]{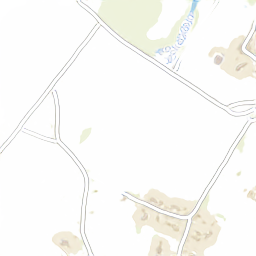}\\
    \includegraphics[width=0.12\linewidth]{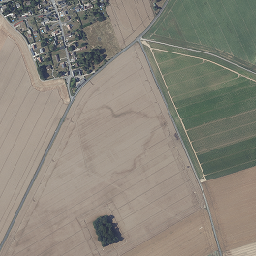}&
    \includegraphics[width=0.12\linewidth]{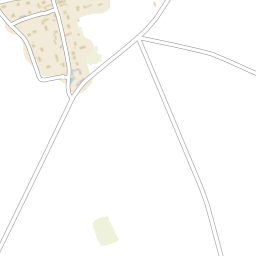}&
    \includegraphics[width=0.12\linewidth]{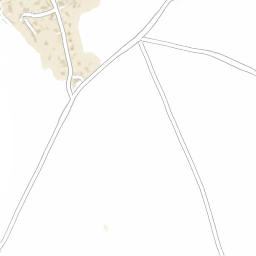}&
    \includegraphics[width=0.12\linewidth]{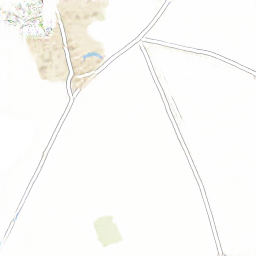}&
    \includegraphics[width=0.12\linewidth]{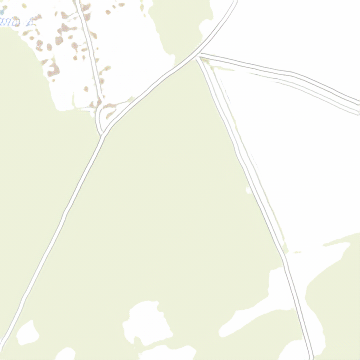}&
    \includegraphics[width=0.12\linewidth]{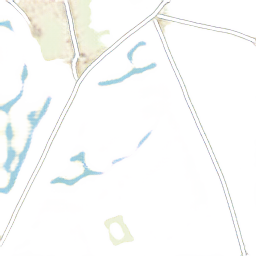}&
    \includegraphics[width=0.12\linewidth]{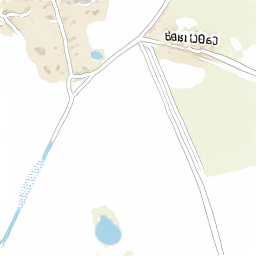}&
    \includegraphics[width=0.12\linewidth]{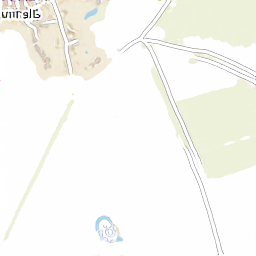}\\
    \includegraphics[width=0.12\linewidth]{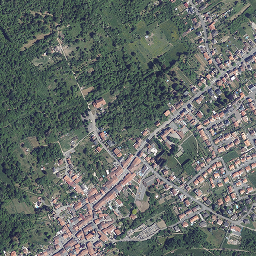}&
    \includegraphics[width=0.12\linewidth]{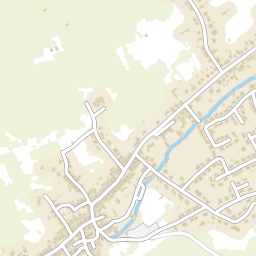}&
    \includegraphics[width=0.12\linewidth]{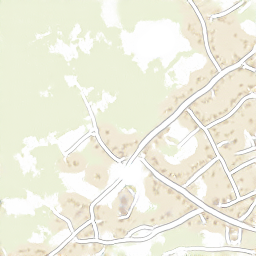}&
    \includegraphics[width=0.12\linewidth]{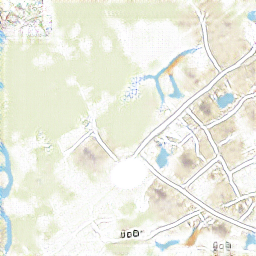}&
    \includegraphics[width=0.12\linewidth]{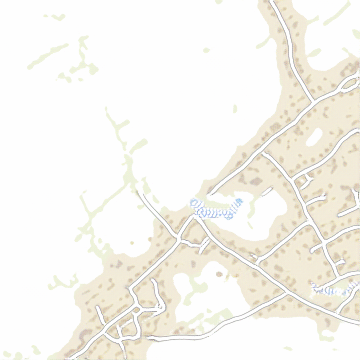}&
    \includegraphics[width=0.12\linewidth]{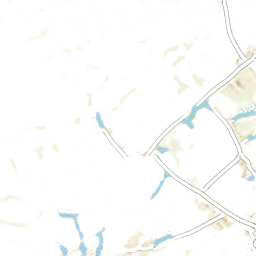}&
    \includegraphics[width=0.12\linewidth]{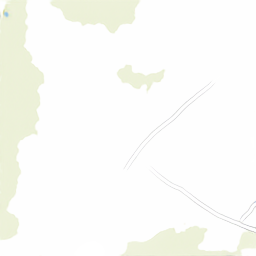}&
    \includegraphics[width=0.12\linewidth]{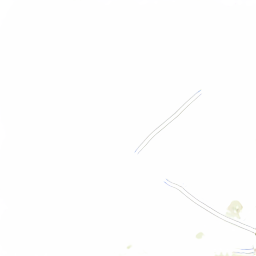}\\
  \end{tabularx}
  \caption{Additional qualitative comparison on \textbf{PlanIGN}.}
  \label{fig:PlanIGN_qualitative_additional}
\end{figure*}

%% file: figures/Brats_qualitative_additional.tex
\begin{figure*}[t]
  \centering
  \setlength\tabcolsep{1pt}
  \begin{tabularx}{\linewidth}{cccccccc}
    Source & Ground Truth & \bf StegoGAN & CycleGAN \cite{CycleGAN2017} & DRIT \cite{Lee2018DRIT} & GcGAN \cite{FuCVPR19-GcGAN} & CUT \cite{park2020cut} & SRUNIT \cite{jia2021srunit}\\
    \includegraphics[angle=-90, width=0.12\linewidth]{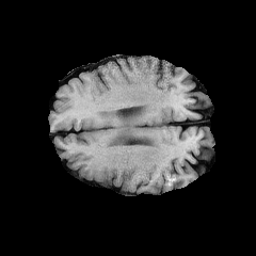}
    &
    \includegraphics[angle=-90, width=0.12\linewidth]{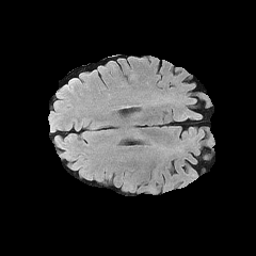}
    &
    \includegraphics[angle=-90, width=0.12\linewidth]{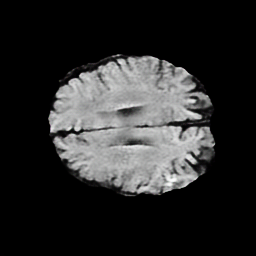}
    &
    \includegraphics[angle=-90, width=0.12\linewidth]{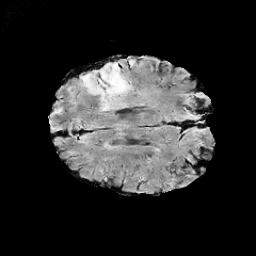}
    &
    \includegraphics[angle=-90, width=0.12\linewidth]{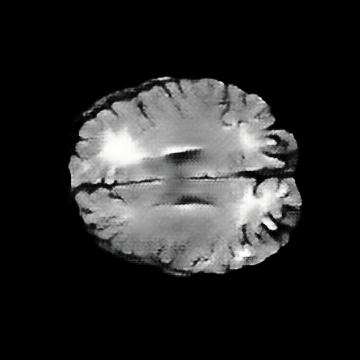}
    &
    \includegraphics[angle=-90, width=0.12\linewidth]{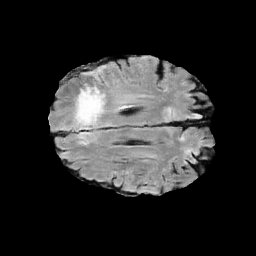}
    &
    \includegraphics[angle=-90, width=0.12\linewidth]{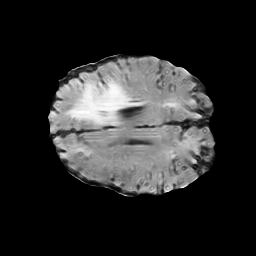}
    &
    \includegraphics[angle=-90, width=0.12\linewidth]{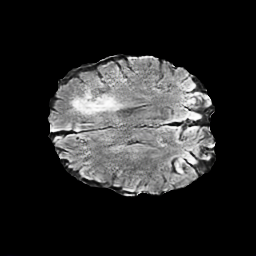}\\
    \includegraphics[angle=-90, width=0.12\linewidth]{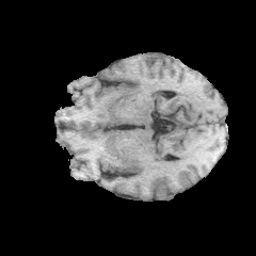}
    &
    \includegraphics[angle=-90, width=0.12\linewidth]{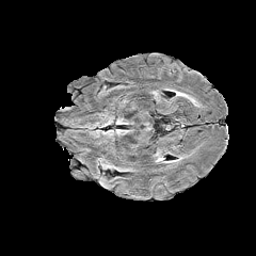}
    &
    \includegraphics[angle=-90, width=0.12\linewidth]{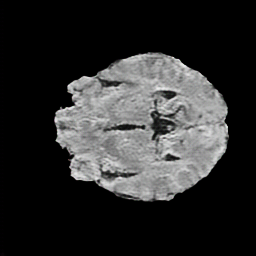}
    &
    \includegraphics[angle=-90, width=0.12\linewidth]{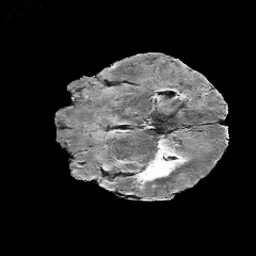}
    &
    \includegraphics[angle=-90, width=0.12\linewidth]{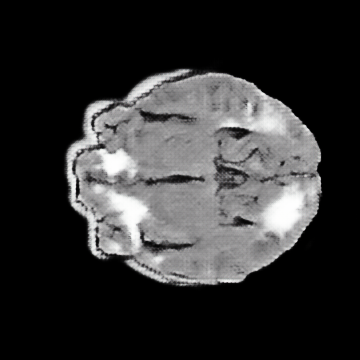}
    &
    \includegraphics[angle=-90, width=0.12\linewidth]{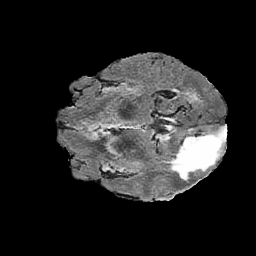}
    &
    \includegraphics[angle=-90, width=0.12\linewidth]{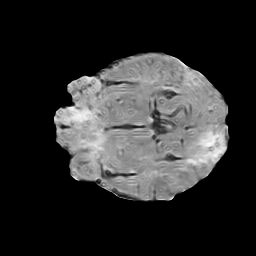}
    &
    \includegraphics[angle=-90, width=0.12\linewidth]{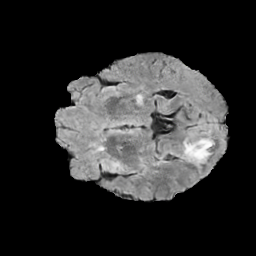}\\
    \includegraphics[angle=-90, width=0.12\linewidth]{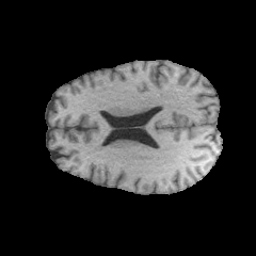}
    &
    \includegraphics[angle=-90, width=0.12\linewidth]{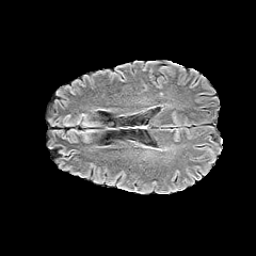}
    &
    \includegraphics[angle=-90, width=0.12\linewidth]{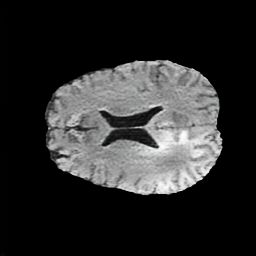}
    &
    \includegraphics[angle=-90, width=0.12\linewidth]{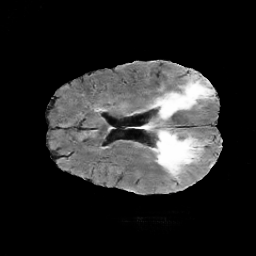}
    &
    \includegraphics[angle=-90, width=0.12\linewidth]{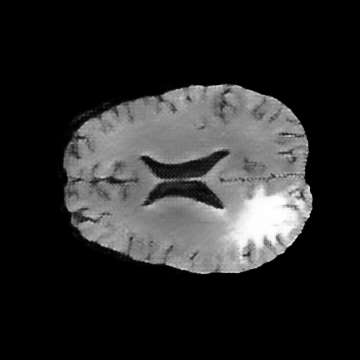}
    &
    \includegraphics[angle=-90, width=0.12\linewidth]{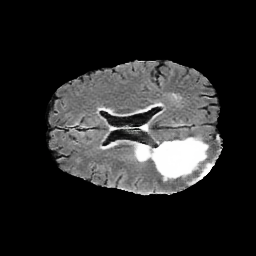}
    &
    \includegraphics[angle=-90, width=0.12\linewidth]{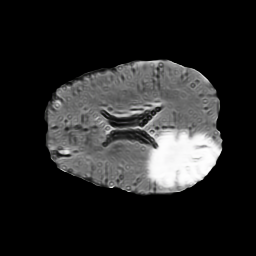}
    &
    \includegraphics[angle=-90, width=0.12\linewidth]{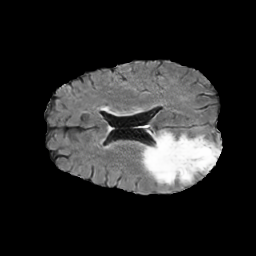}\\    
  \end{tabularx}
  \caption{{Additional qualitative comparison on \textbf{Brats}.} 
  }
  \label{fig:brats_qualitative_additional}
\vspace{-0.5cm}
\end{figure*}